\documentclass{article}

\usepackage{arxiv}

\usepackage[utf8]{inputenc} % allow utf-8 input
\usepackage[T1]{fontenc}    % use 8-bit T1 fonts
\usepackage{hyperref}       % hyperlinks
\usepackage{url}            % simple URL typesetting
\usepackage{booktabs}       % professional-quality tables
\usepackage{amsfonts}       % blackboard math symbols
\usepackage{nicefrac}       % compact symbols for 1/2, etc.
\usepackage{microtype}      % microtypography
\usepackage{lipsum}		% Can be removed after putting your text content
\usepackage{graphicx}

\usepackage{amssymb}
\usepackage{amsthm}
\usepackage{algorithm}
\usepackage{algorithmic}
\usepackage{graphicx}
\usepackage{amsmath}

\title{Pattern Similarity-based Machine Learning Methods for Mid-term Load Forecasting: A Comparative Study}

\author{ \href{https://orcid.org/0000-0002-2285-0327}{\includegraphics[scale=0.06]{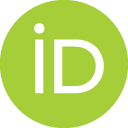}\hspace{1mm}Grzegorz Dudek}
	\\
	Department of Electrical Engineering\\
	Częstochowa University of Technology\\
	Częstochowa, Poland \\
	\texttt{dudek@el.pcz.czest.pl} \\
	\And
	\href{https://orcid.org/0000-0002-2609-811X}{\includegraphics[scale=0.06]{orcid.png}\hspace{1mm}Paweł Pełka} \\
	Department of Electrical Engineering\\
	Częstochowa University of Technology\\
	Częstochowa, Poland \\
	\texttt{p.pelka@el.pcz.czest.pl} \\
}

\begin{document}
\maketitle

\begin{abstract}
	Pattern similarity-based methods are widely used in classification and regression problems. Repeated, similar-shaped cycles observed in seasonal time series encourage to apply these methods for forecasting. In this paper we use the pattern similarity-based methods for forecasting monthly electricity demand expressing annual seasonality. An integral part of the models is the time series representation using patterns of time series sequences. Pattern representation ensures the input and output data unification through trend filtering and variance equalization. Consequently, pattern representation simplifies the forecasting problem and allows us to use models based on pattern similarity. We consider four such models: nearest neighbor model, fuzzy neighborhood model, kernel regression model and general regression neural network. A regression function is constructed by aggregation output patterns with weights dependent on the similarity between input patterns. The advantages of the proposed models are: clear principle of operation, small number of parameters to adjust, fast optimization procedure, good generalization ability, working on the newest data without retraining, robustness to missing input variables, and generating a vector as an output. 
	
	In the experimental part of the work the proposed models were used to forecasting the monthly demand for 35 European countries. The model performances were compared with the performances of the classical models such as ARIMA and exponential smoothing as well as state-of-the-art models such as multilayer perceptron, neuro-fuzzy system and long short-term memory model. The results show high performance of the proposed models which outperform the comparative models in accuracy, simplicity and ease of optimization.  
\end{abstract}

% keywords can be removed

\section{Introduction}
\label{}

Mid-term electrical load forecasting (MTLF) is an essential tool for power system operation and planning. It concerns forecasting monthly electricity demand as well as daily peak load for the next months. The forecast horizon is usually from a week to a year. The mid-term load forecasts are necessary for  maintenance scheduling, fuel reserve planning, hydro-thermal coordination,  electrical energy import/export planing and security assessment. In the deregulated power systems MTLF is a basis for negotiation of forward contracts. The forecast accuracy translates into financial performance of energy companies and other energy market participants.       

This work focuses on the monthly electricity demand forecasting. Time series of monthly electricity demand usually exhibits a trend and annual cycles. The trend is dependent on the dynamics of economic development of a country. Seasonal cycles are related to the climate, weather factors and variability of seasons. Factors which disturb the time series include political decisions, unpredictable economic events, structural breaks \cite{Dog16} and transitory effects from external variables \cite{Apa12}.

Methods of MTLF can be roughly classified to the conditional modeling approach and autonomous modeling approach \cite{Ghi06}. The conditional modeling approach focuses on economic analysis and long-term planning and forecasting of energy policy. Socio-economic conditions that affect energy demand in a given region and population migrations are taken into account. Economic growth is described by economic indicators, which are additional inputs to the forecasting model. Such indicators include \cite{Ghi06}, \cite{Gav01}: gross national product, consumer price index, exchange rates or average wage. In addition, variables describing the power system and network infrastructure are introduced as inputs such as the number and length of transmission lines and the number of high voltage stations. Economic variables have the greatest impact on the trend, while weather variables, due to their seasonal nature, on the periodic behavior of the monthly electricity demand \cite{Gon08}. Examples of conditional modeling approach can be found in \cite{Kan02}, \cite{Bun09} and \cite{Moh18}. In \cite{Kan02} a knowledge-based expert system dedicated for fast developing utility is proposed. It identifies forecasting algorithms and the key variables, electrical and nonelectrical ones, that affect demand forecasts. A set of decision rules relating these variables are then obtained and stored in the knowledge base. Afterwards, the best model that will reflect accurately the typical system behavior over other models is suggested to produce the load forecast. Multiple linear regression and ARIMA models for the monthly peak load forecasing are proposed in  \cite{Bun09}. Inputs of the models are: historical series of electric peak load, weather variables and economic variables such as consumer price index, and industrial index. In \cite{Moh18} the heuristic model was proposed which approximates the relationship between the actual load and four sets of historical data: population, gross national product, consumer price index and temperature. Also the impact of the reserve margin and load diversity factor is considered to obtain the final forecast. 

In autonomous modeling approach, the input variables include only historical loads and weather factors. This approach is more appropriate for stable economies, without rapid changes affecting electricity demand. The selection of weather variables, such as \cite{Ghi06}: atmospheric temperature, humidity, insolation time, wind speed, etc., depends on the local climate and weather conditions \cite{Dov99}. Autonomous models described in \cite{Elk98} use ARIMA and neural networks (NNs) for forecasting monthly peak loads. Input variables include load profiles, weather factors (temperature and humidity) and time index. The model described in \cite{Dov99} uses historical demand and atmospheric temperatures as input variables. Variables expressing the seasons are also introduced. In many cases, autonomous models are simplified using only historical load data as input. Such an approach was used in \cite{Gon08}, where the trend of a series of monthly loads was forecasted only on the basis of loads from the previous twelve months. The seasonal component was modeled by Fourier series. In \cite{Pei11} a fuzzy neural network model was used based only on weather variables (atmospheric pressure, temperature, wind speed, etc.), without taking into account historical loads as input variables.

Another categorization of MTLF methods is based on the forecasting models which can be classical statistical/econometrics models or computational intelligence/machine learning models \cite{Sug11}. Typical examples of the former are ARIMA, linear regression (LR) and exponential smoothing (EST). Implementation of seasonal cycles in the LR models requires additional operations, such as decomposition of the series into individual month series. For nonstationary time series with an irregular periodic trend in \cite{Bar01} a LR model extended with periodic components implemented by the sine functions of different frequencies has been proposed. Another example of using LR for forecasting power systems loads can be found in \cite{AlH05}. The model uses strong daily and yearly correlations to forecast daily load profiles over a period of several weeks to several years. Forecast results are corrected by annual load increases. In \cite{Bun09} the LR and ARIMA models were compared in the task of forecasting monthly peak loads up to 12 months ahead. Models use the same set of input variables including historical peak load data, weather and economic data. In experimental studies, ARIMA proved to be about twice as accurate as LR.

Limited adaptability of the classical MTLF methods and problems with modeling of nonlinear relationships have caused an increase in interest in artificial intelligence and machine learning methods \cite{Gon08}. The most explored machine learning models in MTLF are neural networks. This is due to their attractive features such as nonlinear modeling, learning capabilities, universal approximation property and massive parallelism. In \cite{Gon06} for forecasting a trend of the monthly load time series and seasonal fluctuations two separate NNs are used. In \cite{Chen17} NN on the basis of historical loads and whether variables predicts the future monthly loads. To improve learning capability, NN is learned using gravitational search algorithm and cuckoo optimization algorithm. An example of using Kohonen NN for MTLF can be found in \cite{Gav01}. The authors built 12 forecasting networks for each month of the year. Input vectors contained historical loads and microeconomic indicators. NN proposed in \cite{Dov99} are supported by fuzzy logic. Seasonal variables are defined in the form of trapezoidal indicators of the season. The authors train a set of NNs using regularization techniques to prevent overfitting and aggregate their responses which results in more accurate forecasts. Other examples of machine learning models for MTLF are: \cite{Pei11}, where weighted evolving fuzzy NN for monthly electricity demand forecasting was used, \cite{Ahm19} where NNs, LR and AdaBoost were used for energy forecasting, \cite{Zhao12} where support vector machine was used, and \cite{Bed18} where a long short-term memory network model was used. 

Many of the MTLF methods mentioned above need decomposition of the load time series to deal with a trend and seasonal variations. A typical approach is to decompose the time series into trend, seasonal, and stochastic components. The components expressing less complexity than the original time series can be modeled independently using simpler models. One of the popular tool for decomposition is STL (seasonal and trend decomposition using Loess) filtering procedure based on locally weighted polynomial smoother \cite{The11}. In  \cite{Oli18} STL decomposition was used for monthly data of total electric energy consumption in different developed and developing countries. The times series were forecasted using ARIMA or ETS, and also bootstrap aggregating method was used to improve the model accuracy. Another method of decomposition is a wavelet transform which splits up the load time series into subseries in the wavelet domain. A low frequency component called an approximation expresses a trend while high-frequency components called details express cyclical variations \cite{Ben06}. As an alternative of wavelet decomposition a Fourier transform can be used which decomposes function of time into its constituent frequencies. For example, in \cite{Gon08} the load time series was split into two components: one describing the trend and the other the fluctuations. Then the fluctuation series was expressed by several Fourier series of different frequencies. Yet another method of load time series decomposition is empirical mode decomposition which breaks down the time series into so-called intrinsic mode functions. Such decomposition was used in \cite{Qiu17}, where to model each of the extracted intrinsic mode functions deep belief network was used.

As an alternative of the classical and state-of-the-art methods for MTLF, in this work we describe pattern similarity-based forecasting methods (PSFMs). Similarity-based learning is a practical learning framework as a generalization of the minimal distance methods. It is a basis of many machine learning and pattern recognition methods used for classification, clustering and regression. Repeated, similar-shaped cycles observed in seasonal time series encourage us to apply these methods also for forecasting. To do so, first we define patterns expressing preprocessed repetitive sequences in a time series. Pattern representation ensures the input and output data unification through trend filtering and variance equalization. Consequently, no decomposition of the time series is needed. Due to pattern representation the relationship between input and output data is simplified and the forecasting problem can be solved using simple models. We consider four such models: nearest neighbor model, fuzzy neighborhood model, kernel regression model and general regression neural network. A regression function is constructed by aggregation output patterns with weights dependent on the similarity between input patterns. So, the principle of operation is very clear which is a big advantage of these models compared to other forecasting models that are often black boxes. Other advantages are: small number of parameters to adjust which implies fast optimization procedure, good generalization ability which can be controlled by the model parameters, working on the newest data without retraining, robustness to missing input variables, and generating a vector as an output. Unlike the state-of-the-art machine learning methods, such as neural networks and deep learning, PSFMs do not suffer with excessive tuning and training burden. 

The PSFMs belong to the group of autonomous modeling approach, where only the historical load data are used as inputs. Not including other inputs such as whether and economic factors can raise objections but we have to remember that these exogenous variables are usually not available and have to be forecasted. This is a major challenge in all MTLF approaches with exogenous inputs. Many researchers use statistical measures such as correlation coefficients, personal experience and intuition to assess the validity, effectiveness and contribution of such exogenous variables to energy and load forecasting \cite{Ghi06}. This can lead to low accuracy forecasts of whether and economic variables, and consequently, to large load forecast errors.    

This paper is organized as follows. Section 2 describes monthly electricity demand time series and their representation using patterns. The idea and framework of PSFM as well as the four forecasting models are presented in Section 3. Section 4 describes the experimental framework used to evaluate the performance of the proposed and comparative models.  Finally, Section 5 presents our conclusions.

\section{Time Series and their Representation}
\label{}
\begin{figure}
	\centering
	\includegraphics[width=0.49\textwidth]{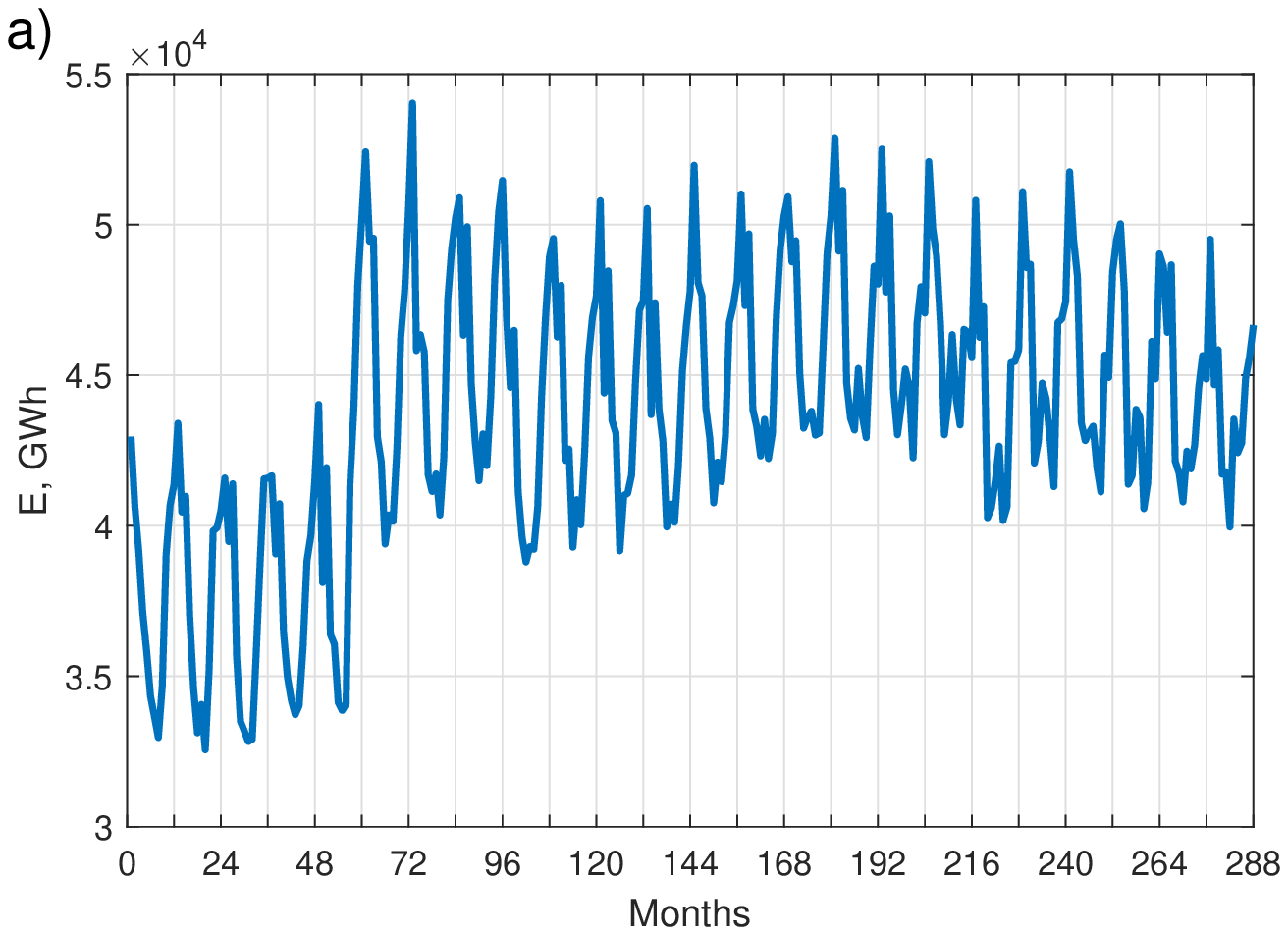}
	\includegraphics[width=0.49\textwidth]{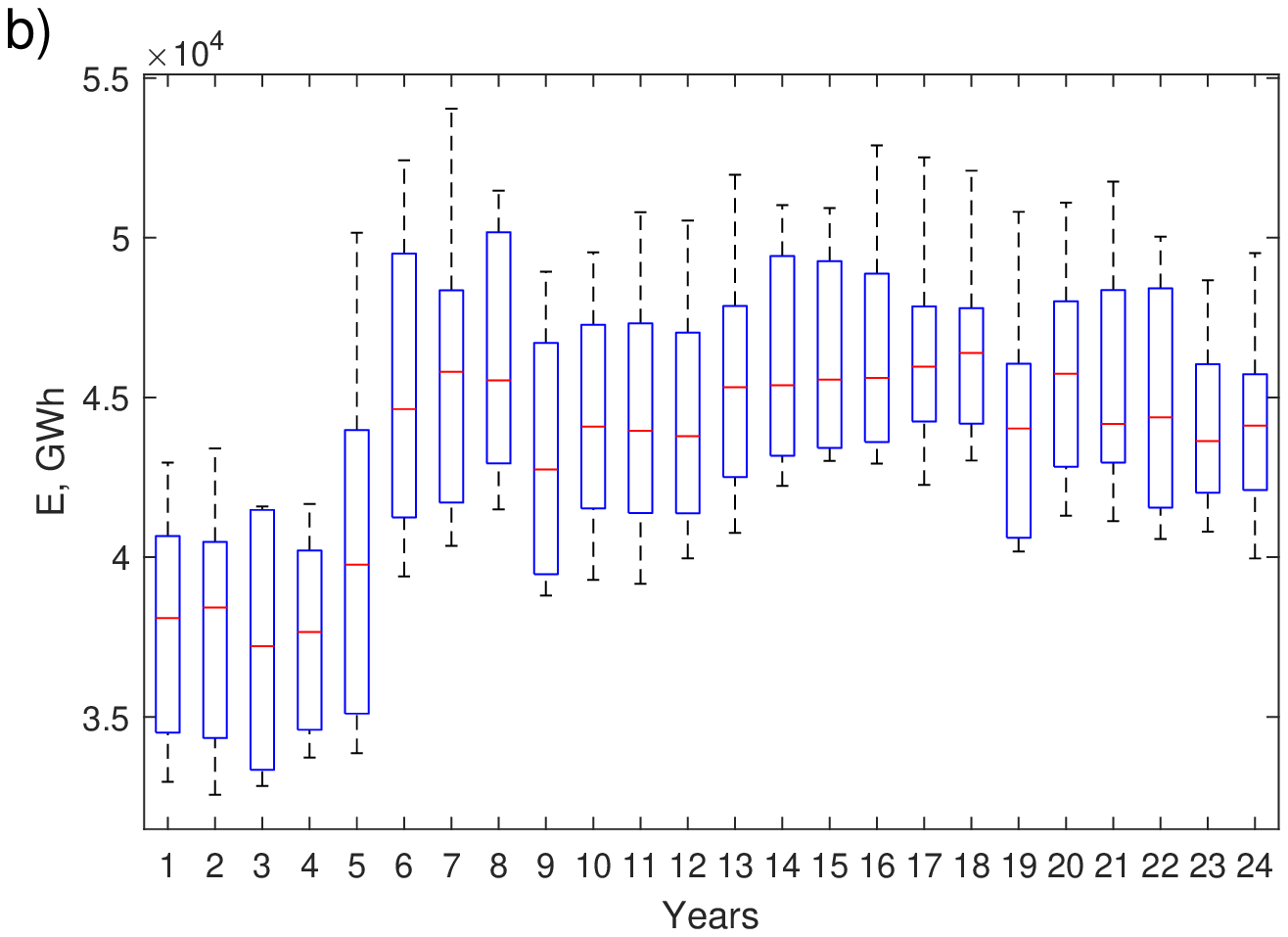}
	\caption{Monthly electricity demand time series for Germany (a) and its dispersion in successive years (b).} 
	\label{fig1}
\end{figure}

A monthly electricity demand time series exhibits a trend, annual cycles and random component. Fig. \ref{fig1} demonstrates an example of such a time series. As you can see in this figure we can observe nonlinear trend, jump at the end of the fifth cycle, and changing annual pattern over the years. Also the dispersion of the annual cycles changes significantly over time (from $\sigma = 2610$ to $5588$ MWh).

One of the main issues in building forecasting models is how the time series should be represented to get the highest performance of the model. Input and output variables should be defined on the basis of the original time series. These definitions affect significantly the results. In this work we use input and output variables as patterns of the fragments of monthly load time series. By a pattern we mean a vector with components that are calculated using some function of actual time series elements. For multiple seasonal time series this function can filter out a trend and seasonal fluctuations and so simplify the data and relationships between them \cite{dud15a}. Thus the forecasting model working on patterns can be less complex. 

An input pattern $\mathbf{x}_i = [x_{i,1} x_{i,2} … x_{i,n}]^T$ of length $n$ is a vector of predictors representing a sequence $ X_i = \{E_{i-n+1}, E_{i-n+2},…, E_i \} $ of $n$ successive time series elements $E_i$ (monthly electricity demands) preceding the forecasted period. A function transforming time series elements into patterns depends on the time series character such as seasonalities, variance and trend. Some definitions of the function mapping original time series into patterns $\mathbf{x}$ are:

\begin{equation}\label{eq1}
x_{i,t} = E_{i-n+t}
\end{equation}
\begin{equation}\label{eq2}
x_{i,t} = E_{i-n+t}-\overline{E}_i
\end{equation}
\begin{equation}\label{eq3}
x_{i,t} = \frac{E_{i-n+t}}{\overline{E}_i}
\end{equation}
\begin{equation}\label{eq4}
x_{i,t} = \frac{E_{i-n+t}-\overline{E}_i}{D_i}
\end{equation}
where $t = 1, 2, ..., n$, $\overline{E}_i$ is a mean of sequence $X_i$, and $D_i = \sqrt{\sum_{j=1}^{n} (E_{i-n+j}-\overline{E}_i)^2}$ is a measure of its dispersion.

Definition \eqref{eq1} just copy sequence $X_i$ into x-pattern without transformation. The pattern components defined using \eqref{eq2} are the differences in demand of a given month and an average demand of the sequence $X_i$. A quotient of this two quantities is expressed by \eqref{eq3}. An x-pattern defined by \eqref{eq4} is a normalized version of a vector composed of the elements of $X_i$, i.e. $[E_{i-n+1} E_{i-n+2}  … E_i]^T$. So, the original time series sequences having different mean and dispersion (see Fig. \ref{fig1} (b)) are unified and after normalization they are represented by x-patterns which all have zero mean, the same variance and also unity length. 

\begin{figure}
	\centering
	\includegraphics[width=0.8\textwidth]{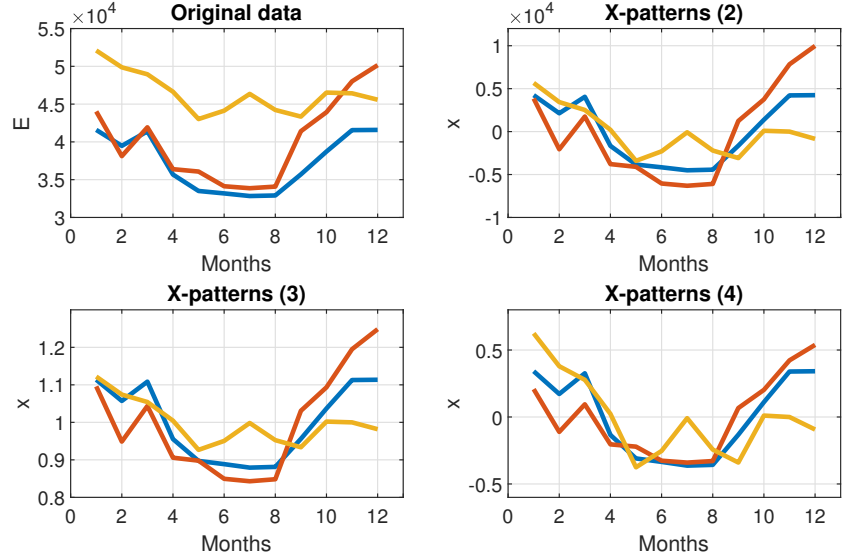}
	\caption{Three annual periods and their x-patterns (time series for Germany).} 
	\label{fig2}
\end{figure}

In Fig. \ref{fig2} one-year sequences $X_i$ are shown and their x-patterns defined using \eqref{eq2}-\eqref{eq4}. Note that all x-pattern definitions boil down the mean of all patterns to the same value (0 or 1) and additionally definition \eqref{eq4} boils down the variance of all patterns to the same value. So, a trend was filtered out and patterns differs only in shape.  

An output pattern $\mathbf{y}_i = [y_{i,1} y_{i,2} … y_{i,m}]^T$ represents a forecasted sequence of length $m$: $ Y_i = \{ E_{i+\tau}, E_{i+\tau+1},…, E_{i+\tau+m-1}\} $, where $\tau$ is a forecast horizon in months. The output patterns are defined similarly to the input patterns:   

\begin{equation}\label{eq5}
y_{i,t} = E_{i+\tau+t-1}
\end{equation}
\begin{equation}\label{eq6}
y_{i,t} = E_{i+\tau+t-1}-\overline{E}_i^*
\end{equation}
\begin{equation}\label{eq7}
y_{i,t} = \frac{E_{i+\tau+t-1}}{\overline{E}_i^*}
\end{equation}
\begin{equation}\label{eq8}
y_{i,t} = \frac{E_{i+\tau+t-1}-\overline{E}_i^*}{D_i^*}
\end{equation}
where $t = 1, 2, ..., m$, and $\overline{E}_i^*$ and $D_i^*$ are coding variables.

The coding variables, $\overline{E}_i^*$ and $D_i^*$, are the mean and dispersion of the forecasted sequence $Y_i$, respectively. 
They both are known for the historical time series, so the training y-patterns can be prepared using them and the model can be trained. 
After learning, for a new query x-pattern, the model produces the forecasted y-pattern, $\widehat{\mathbf{y}}$. The forecasts of the monthly electricity demands in period $Y_i$ are calculated from the forecasted y-pattern using transformed equations $\eqref{eq5}-\eqref{eq8}$ (this is called decoding). For example, when y-patterns are defined using \eqref{eq8}, a decoding equation is of the form:

\begin{equation}\label{eq9}
\widehat{E}_{i+\tau+t-1} = \widehat{y}_{i,t}D_i^* + \overline{E}_i^*, \quad t = 1, 2, ..., m
\end{equation}  

In this equation the coding variables are not known, because they are the mean and dispersion of the future sequence $Y_i$, which is just forecasted. In this case the coding variables are predicted from their historical values. In the experimental part of the work the coding variables are predicted using ARIMA and exponential smoothing (ETS). 

To avoid forecasting the coding variables we use another approach. Instead of using the mean and dispersion of the forecasted sequence $Y_i$ as coding variables, we introduce in \eqref{eq6}-\eqref{eq8} as coding variables the mean and dispersion of sequence $X_i$, i.e. $\overline{E}_i^*=\overline{E}_i$, $D_i^*=D_i$. Although this approach does not guarantee that all y-patterns have the same mean value, it unifies output data taking into account the current process variability, expressed by mean $\overline{E}_i$ and dispersion $D_i$. When the model returns the forecasted y-patterns, the forecast of the monthly demands are calculated from \eqref{eq9} using known coding variables for the historical sequence $X_i$.

A training set for learning forecasting models is composed of pairs of corresponding x- and y-patterns: $\Phi = \{ (\mathbf{x}_i, \mathbf{y}_i) | \mathbf{x}_i \in \mathbb{R}^n, \mathbf{y}_i \in \mathbb{R}^m, i = 1, 2, ..., N \} $. The model learns the mapping x-patterns $\rightarrow$ y-patterns. It generates a forecast of a y-pattern, $\widehat{\mathbf{y}}$, for a query pattern $\mathbf{x}$. The forecasts of the monthly electricity demands in period $Y_i$ are calculated from the forecasted y-pattern using transformed equations \eqref{eq5}-\eqref{eq8}.

Note that when using pattern approach, the forecasting model works on patterns expressing shapes of the time series sequences. In the first step of this approach the trend and dispersion (and also additional seasonal variation in the time series with multiple seasonal cycles) are filtered out. Then the model forecasts the unified data, i.e. y-patterns on the basis of x-patterns. Finally, in the decoding step, current trend and dispersion is introduced to the forecasted y-pattern to get the forecasted monthly demand.

\section{Pattern Similarity-based Forecasting Models}
\label{}

\subsection{The Framework of the PSFM}
\label{}

Similarity-based  methods are very popular in the field of machine learning and pattern recognition \cite{duch}, \cite{chen09}. They estimate the class label or the function value for the query sample based on the similarities between this sample and a set of training samples using some similarity function defined for any pair of samples. The proposed forecastng methods can be classified as memory-based approximation methods using
analogies between preprocessed sequences of the time series (patterns). It is assumed that a time series behavior in the future can be deduced from its behavior in similar conditions in the past. This assumption can be expressed in the pattern representation context as follows \cite{dud15a}: If the query pattern $\mathbf{x}$ is similar to pattern $\mathbf{x}_i$ form the history, then the forecasted pattern $\mathbf{y}$ will be similar to pattern $\mathbf{y}_i$ (paired with $\mathbf{x}_i$). This assumption allows us to predict the y-pattern on the basis of known patterns $\mathbf{x}$,  $\mathbf{x}_i$ and $\mathbf{y}_i$. Usually we select several patterns $\mathbf{x}_i$ and aggregate patterns $\mathbf{y}_i$ paired with them to get the forecasted y-pattern. 

The above assumption underlying PSFMs should be verified for a given time series. We analyse relationship between similarities of x-patterns and similarities of paired with them y-patterns. To do so, we define two random variables: similarity between $\mathbf{x}_i$ and $\mathbf{x}_j $, $s(\mathbf{x}_i,\mathbf{x}_j)$, and similarity between $\mathbf{y}_i$ and $\mathbf{y}_j $, $s(\mathbf{y}_i,\mathbf{y}_j)$, where $i,j=1,2,...,N, i \neq j$. Instead of similarity measure we can use a distance measure between patterns as random variables. All pairs of these random variables form the sample. To show the stochastic interdependence of the random variables the null hypothesis is formulated: The observed differences in numbers of occurrence of the sample elements in the specified categories of the random variables are caused by a random nature of the sample. This hypothesis is verified using the chi-squared test based on a contingency table showing the joint empirical distribution of the random variables. High value of the $\chi^2$ statistic, above the critical value, rejects the null hypothesis in favor of the alternative hypothesis, which justifies the use of PSFMs. 

In this study similarity between patterns, which are real-valued vectors, is measured using Euclidean distance. Other measures are also possible such as: Pearson’s correlation coefficient, Tanimoto coefficient, other Minkowski distances or dot product for normalized vectors.  

\begin{figure}
	\centering
	\includegraphics[width=1\textwidth]{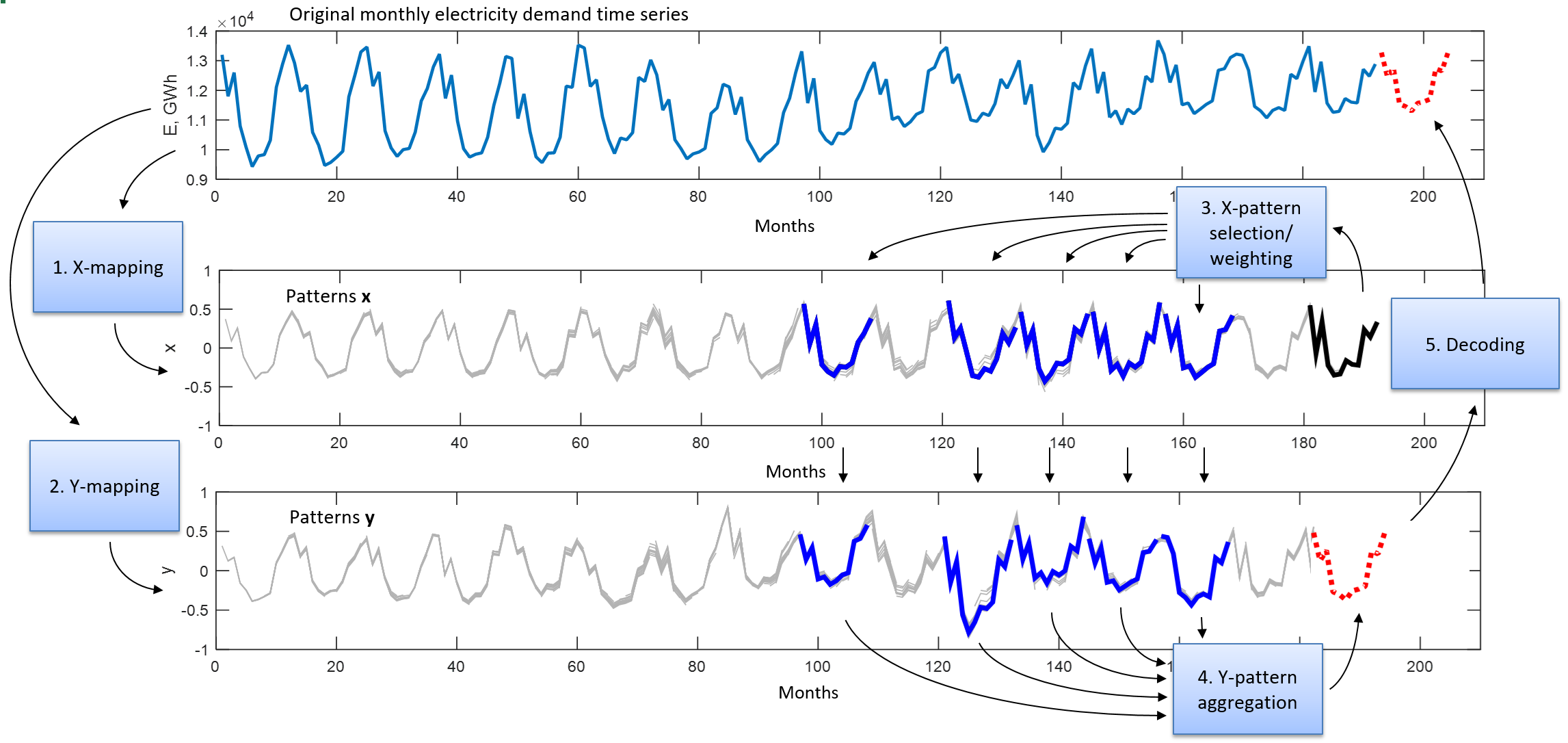}
	\caption{Idea of the pattern similarity-based forecasting.} 
	\label{fig3}
\end{figure}

The pattern similarity-based forecasting procedure is depicted in Fig. \ref{fig3} and can be summarized in the following steps \cite{dud15a}:

\begin{enumerate}
	\item Mapping the original time series sequences into x- and y-patterns.
	\item Selection of the training x-patterns similar to the query pattern $\mathbf{x}$.
	\item Aggregation of the y-patterns paired with the similar x-patterns to get the forecasted pattern $\widehat{\mathbf{y}}$.
	\item Decoding pattern $\widehat{\mathbf{y}}$ to get the forecasted time series sequence $Y$.
\end{enumerate} 

During aggregation of the y-patterns (step 3) we use the weights for them which are dependent on the similarity between a query pattern $\mathbf{x}$ and the training x-patterns. In this case, the regression model mapping x-patterns into y-patterns is of the form:

\begin{equation}\label{eq10}
m(\mathbf{x})= \sum\limits_{i=1}^N w(\mathbf{x}, \mathbf{x}_i)\mathbf{y}_i
\end{equation} 
where $\sum_{i=1}^N w(\mathbf{x}, \mathbf{x}_i)=1$, $w(.,.)$ is a weighting function.

Note that $m(.)$ is a vector-valued function returning the whole y-pattern. It is nonlinear function if $w(.,.)$ maps $\mathbf{x}$ nonlinearly. Different definitions of $w(.,.)$ are presented below where the PSFMs are specified.

\subsection{Nearest Neighbor Models}
\label{}

The simplest PSFM is a $k$-nearest neighbor regression model. It estimates $m(.)$ as the average of the y-patterns in a varying neighborhood of query pattern $\mathbf{x}$. The neighborhood is defined as a set of $k$ nearest neighbors of $\mathbf{x}$ in the training set. The regression function is as follows:

\begin{equation}\label{eq11}
m(\mathbf{x})= \sum\limits_{i \in \Omega_k(\mathbf{x})} w(\mathbf{x}, \mathbf{x}_i) \mathbf{y}_i
\end{equation} 
where $\Omega_k(\mathbf{x})$ is a set of indices of $k$ nearest neighbors of $\mathbf{x}$ in $\Phi$ and $w(\mathbf{x}, \mathbf{x}_i)=1/k$ for each $\mathbf{x}$ and $\mathbf{x}_i$.

Note that in this model the weights of $\mathbf{y}_i$ are all equal to $1/k$. Function \eqref{eq11} is a step function. The number of nearest neighbors $k$ controls the smoothness of the estimator. For $k=1$ the regression function is exactly fitted to the training points. Increasing $k$ causes an increase in bias and decrease in variance of the estimator. To get rid of jumps of the regression function and made it more smooth, we can introduce the weighting function that gives greater weights for closer neighbors and lower weights for distant neighbors. Weighting functions are usually dependent on the distance between a query pattern an its nearest neighbors. They are monotonically decreasing, reach the maximum value in zero, and the minimum value (nonnegative) for the $k$-th nearest neighbor. Some weighting function propositions can be found in \cite{Atk97}. In this work we use the weighting function of the form  \cite{dud15b}: 

\begin{equation}\label{eq12}
w(\mathbf{x}, \mathbf{x}_i)=\frac{v(\mathbf{x}, \mathbf{x}_i)}{\sum\limits_{j \in \Omega_k(\mathbf{x})}v(\mathbf{x}, \mathbf{x}_j)}
\end{equation}
\begin{equation}\label{eq13}
v(\mathbf{x}, \mathbf{x}_i)=\rho \left( \frac{1-d(\mathbf{x},\mathbf{x}_i)/d(\mathbf{x},\mathbf{x}^k)}{1+\gamma d(\mathbf{x},\mathbf{x}_i)/d(\mathbf{x},\mathbf{x}^k)} -1\right) +1
\end{equation}
where $\mathbf{x}^k$  is the $k$-th nearest neighbor of $\mathbf{x}$ in $\Phi$, $d(\mathbf{x},\mathbf{x}_i)$ is a Euclidean distance between $\mathbf{x}$ and its $i$-th nearest neighbor in $\Phi$, $\rho \in [0, 1]$ is a parameter deciding about the differentiation of weights, and $\gamma \geq -1$ is a parameter deciding about the convexity of the weighting function.

\begin{figure}
	\centering
	\includegraphics[width=0.7\textwidth]{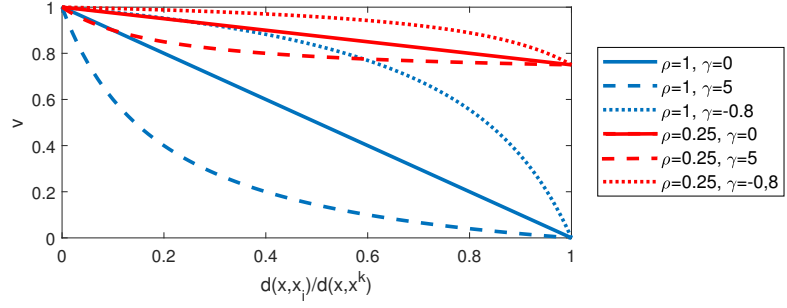}
	\caption{The weighting function for $k$-NN.} 
	\label{fig4}
\end{figure}

The weighting function \eqref{eq13} is shown in Fig. \ref{fig4}. The interval of weights $v$ is $[1-\rho,1]$. So, for $\rho=1$ the weights are the most diverse and for $\rho=0$ they are all equal. In the latter case we get $w(\mathbf{x}, \mathbf{x}_i)=1/k$. For $\gamma = 0$ the weighting function is linear. For $\gamma > 0$ we get convex function and for $\gamma < 0$ we get concave function. The three model parameters, $k, \rho$ and $\gamma$, allows us to control flexibly its features. 

\subsection{Fuzzy Neighborhood Model}
\label{}

\begin{figure}
	\centering
	\includegraphics[width=0.7\textwidth]{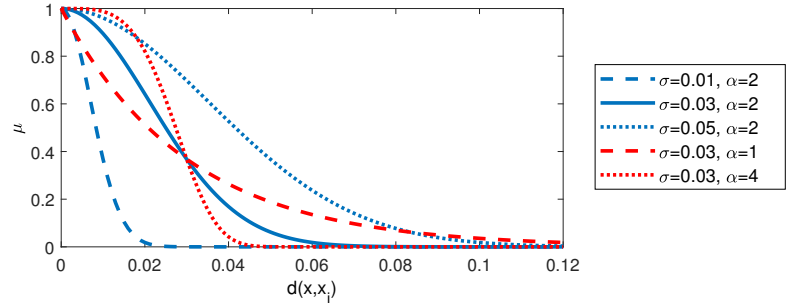}
	\caption{The membership function for FNM.} 
	\label{fig5}
\end{figure}

In the $k$-NN model the regression surface is built using $k$ training patterns. Fuzzy neighborhood model (FNM) takes into account all training patterns when constructing the regression surface \cite{Pel17}. In this case not only $k$ patterns belong to the query pattern neighborhood but all training patterns belong to it, with a different membership degree. The membership function is dependent on the distance between the query pattern $\mathbf{x}$ and the training pattern $\mathbf{x}_i$ as follows:

\begin{equation}\label{eq14}
\mu (\mathbf{x},\mathbf{x}_i)=\exp \left(-\left( \frac{d(\mathbf{x},\mathbf{x}_i)}{\sigma}
\right)^\alpha \right) 
\end{equation}
where $\sigma$ and $\alpha$ are parameters deciding about the membership function shape (see Fig. \ref{fig5}). 

The weighting function in FNM is of the form:

\begin{equation}\label{eq15}
w(\mathbf{x}, \mathbf{x}_i)=\frac{\mu (\mathbf{x},\mathbf{x}_i)}{\sum\limits_{j = 1}^N  \mu(\mathbf{x},\mathbf{x}_j)}
\end{equation}

Membership function \eqref{eq15} is a Gaussian-type function. Other types are also possible, e.g a Cauchy-type function with fatter tail, which gives greater weights for more distant patterns. The model parameters, $\sigma$ and $\alpha$, shape the membership function and thus control the properties of the estimator. For wider membership functions the model tends to increase bias and decrease variance.

\subsection{Kernel Regression Model}
\label{}

Kernel regression belongs to the same category of nonparametric methods as $k$-NN and FNM. It models a nonlinear relationship between a pair of random variables $\mathbf{x}$ and $\mathbf{y}$. The Nadaraya–Watson estimator (N-WE) is the most popular representative of this group. N-WE estimates regression function $m(.)$ as a locally weighted average, using in \eqref{eq10} a kernel $K_h$ as a weighting function \cite{Har04}: 

\begin{equation}\label{eq16}
w(\mathbf{x}, \mathbf{x}_i)=\frac{K_h (\mathbf{x}-\mathbf{x}_i)}{\sum\limits_{j = 1}^N  K_h(\mathbf{x}-\mathbf{x}_j)}
\end{equation}

A kernel function is centered at the data point $\mathbf{x}_i$ and gives the highest value when the distance between this point and the query point $\mathbf{x}_i$ is zero. The kernel function falls with the distance at a speed dependent on the smoothing parameter (or bandwidth) $h$. 

When the input variable is multidimensional, the kernel has a product form. In such a case, for a normal kernel, which is often used in practice, the weighting function is defined as \cite{dud15b}, \cite{Pel17b}:

\begin{equation}\label{eq17}
w(\mathbf{x}, \mathbf{x}_i)=\frac{\exp \left(-\sum\limits_{t = 1}^n 
	\frac{(x_t-x_{i,t})^2}{2h_t^2} \right)}
{\sum\limits_{j = 1}^N \exp \left(-\sum\limits_{t = 1}^n 
	\frac{(x_t-x_{j,t})^2}{2h_t^2} \right)} 
\end{equation}
where $h_t$ is a bandwidth for the $t$-th dimension.

In N-WE the bandwidth has a prominent effect on the estimator shape, whereas the kernel is clearly less important. Note that in \eqref{eq17} we define the bandwidths individually for each dimension. This gives more flexible estimator which allows us to control the influence of each input on the resulting fitted surface. The bandwidths decide about the bias-variance tradeoff of the estimator. For their too low values the estimator is undersmoothed, while for their too large values it is oversmoothed.

\subsection{General Regression Neural Network Model}
\label{}

General Regression Neural Network (GRNN) was proposed in \cite{Spe91} as a variant of radial basis function NN. It is a memory-based network where each neuron correspond to the one training x-pattern. GRNN provides smooth approximation of a target function even with sparse data in a multidimensional space. Due to one pass learning it learns very fast when comparing to other NNs. Other advantages of GRNN are: easy tuning, highly parallel structure and smooth approximation of a target function even with sparse data in a multidimensional space.

\begin{figure}
	\centering
	\includegraphics[width=0.5\textwidth]{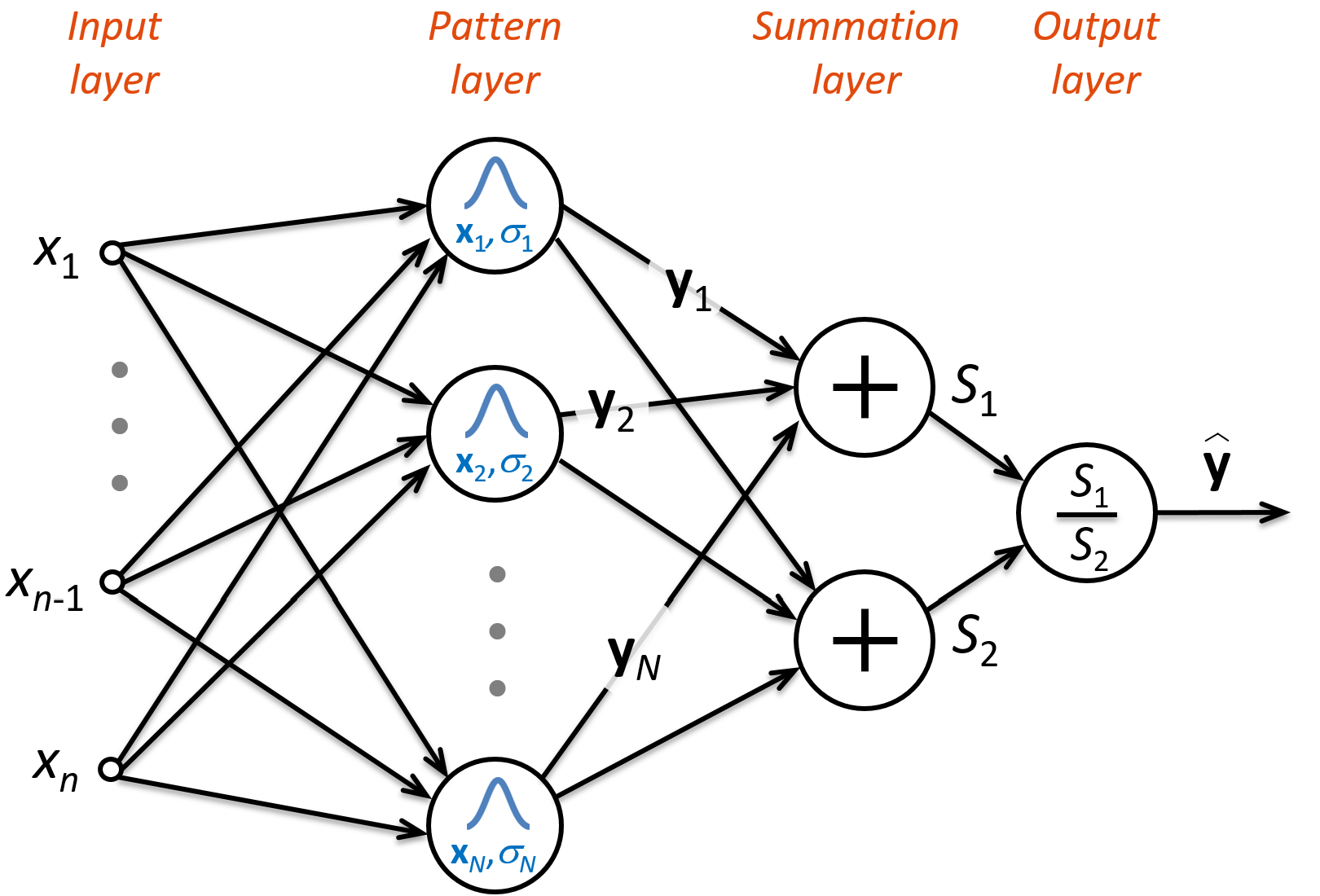}
	\caption{General regression neural network.} 
	\label{fig6}
\end{figure}

As shown in Fig. \ref{fig6}, GRNN is composed of four layers: input, pattern (radial basis layer), summation and output layer. The pattern layer transform inputs nonlinearly using Gaussian activation functions of the form:

\begin{equation}\label{eq18}
G(\mathbf{x},\mathbf{x}_i)=\exp \left(-\frac{\| \mathbf{x}-\mathbf{x}_i) \| ^2}{\sigma_i^2} \right)
\end{equation}
where $\|.\|$ is a Euclidean norm and $\sigma_i$ is a bandwidth for the $i$-th pattern.

The Gaussian functions are centered at different training patterns $\mathbf{x}_i$. The neuron output expresses
similarity between the query pattern and the $i$-th training pattern. This output is treated as the weight of the $i$-th y-pattern. So the pattern layer maps the $n$-dimensional input space into $N$-dimensional space of similarity. The weighting function implemented in GRNN is defined as \cite{Pel19}:

\begin{equation}\label{eq19}
w(\mathbf{x}, \mathbf{x}_i)=\frac{G(\mathbf{x},\mathbf{x}_i)}{\sum\limits_{j = 1}^N  G(\mathbf{x},\mathbf{x}_j)}
\end{equation}

The performance of GRNN is related with bandwidths $\sigma_i$ governing the smoothness of the regression function \eqref{eq10}. Note that in the GRNN model each neuron has its own bandwidth $\sigma_i$. This allow us to control flexibly the weight of the $i$-th y-pattern individually. As in the case of other PSFMs presented above, selection of the optimal values of bandwidths is a key problem in GRNN.   

\section{Simulation Studies}
\label{SS}

\begin{figure}
	\centering
	\includegraphics[width=0.49\textwidth]{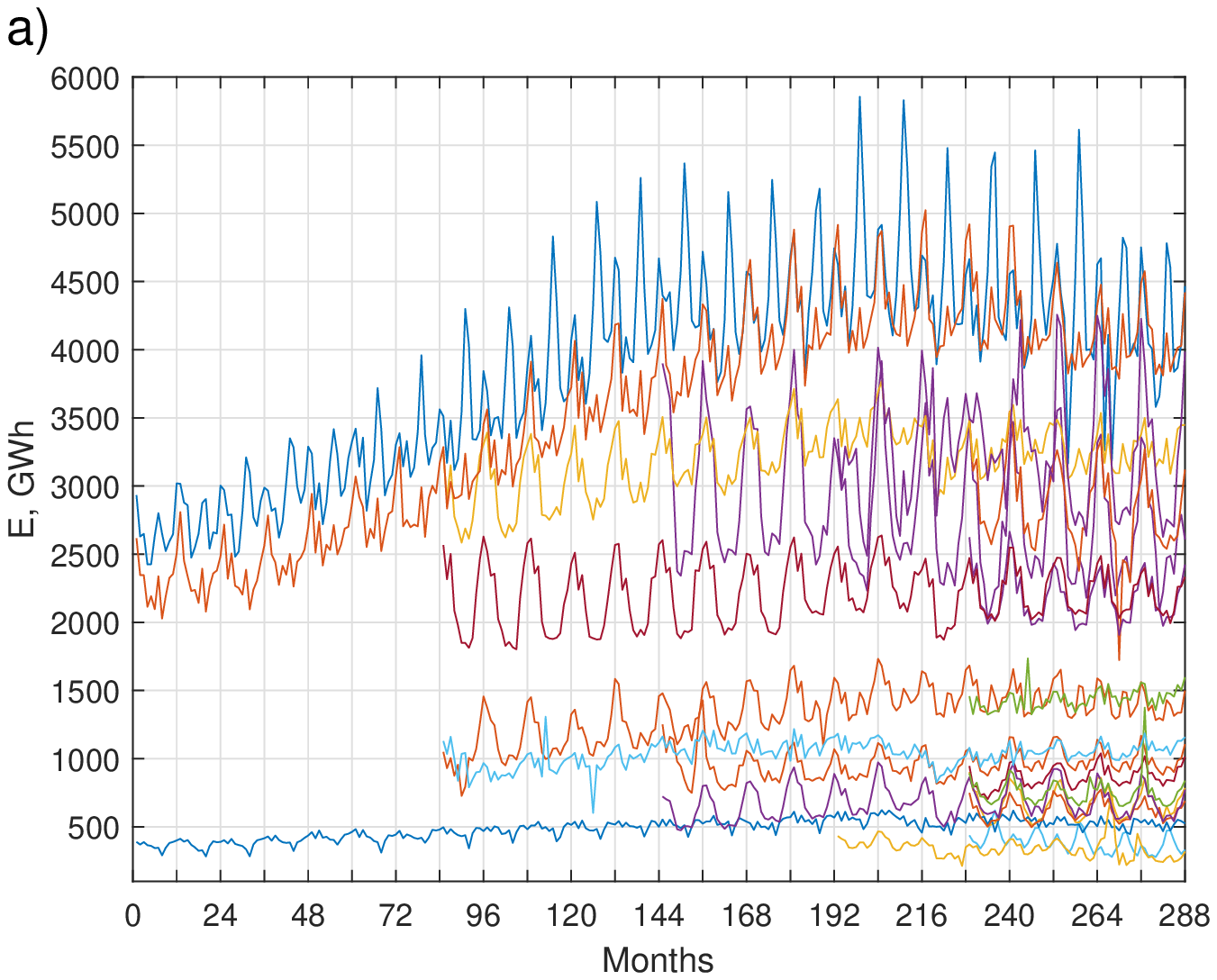}
	\includegraphics[width=0.49\textwidth]{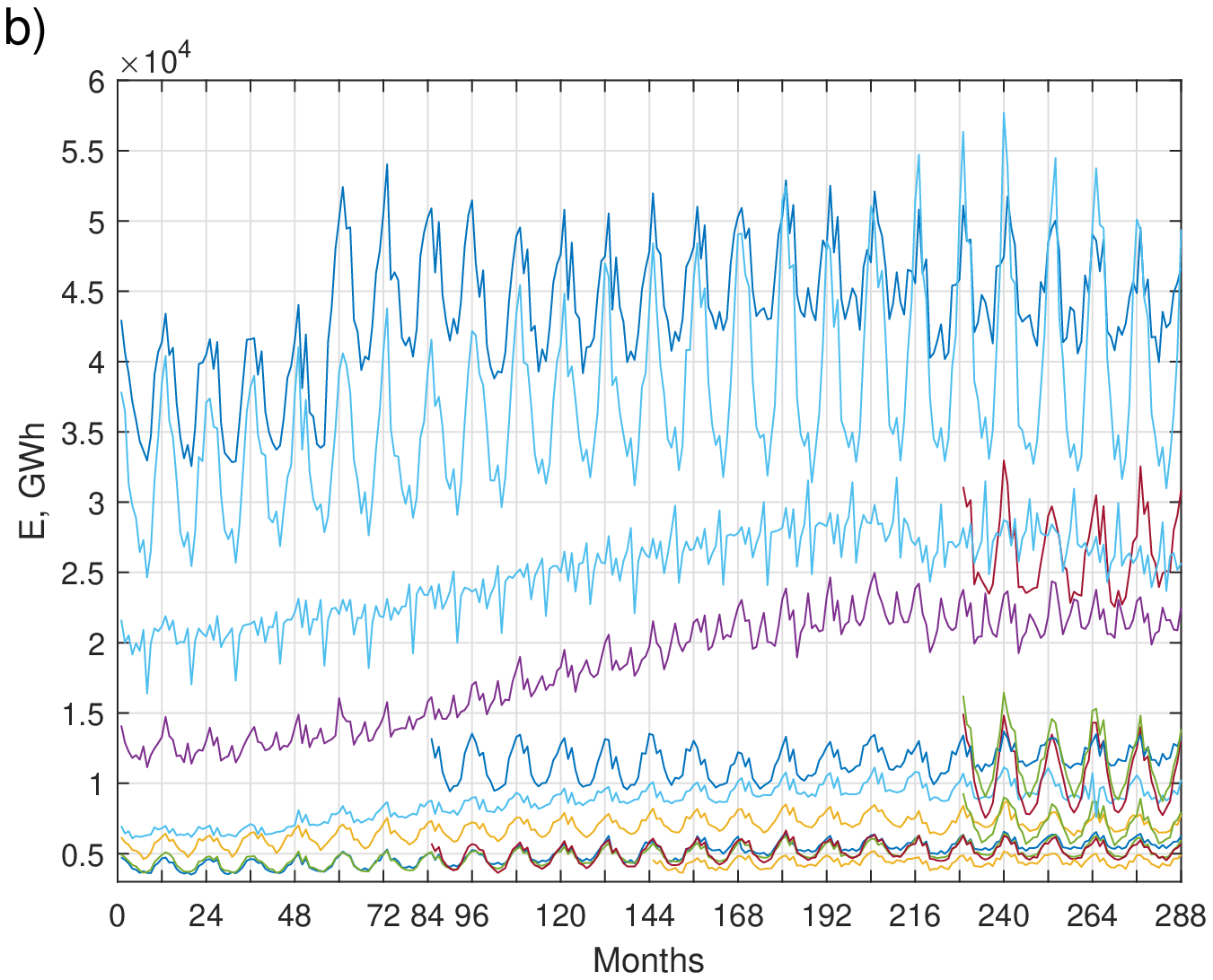}
	\caption{Monthly electricity demand time series for 35 European countries.} 
	\label{fig7}
\end{figure}

In this section we compare PSFMs on the mid-term load forecasting problem using real-world data: monthly electricity demand time series for 35 European countries. The data are taken from the publicly available ENTSO-E repository (www.entsoe.eu). The longest time series cover the time period from 1991 to 2014 (11 countries out of 35). Others are shorter: 17 years (6 countries), 12 years (4 countries), 8 years (2 countries), and 5 years (12 countries). A goal is to predict twelve monthly demands for 2014 using historical data. The 35 time series are shown in Fig. \ref{fig7}. As can be seen from this figure, the time series have different levels, trends, variations and yearly shapes.  

%\subsection{The Data}
%\label{} 

\subsection{Verification of the PSFM assumption}
\label{} 

In Section 3.1 the assumption underlying the PSFMs was stated. To confirm this assumption the null hypothesis was formulated. It is verified for each country using the chi-squared test based on a contingency table showing the joint empirical distribution of the random variables which are Euclidean distances between patterns: $d(\mathbf{x}_i,\mathbf{x}_j)$ and $d(\mathbf{y}_i,\mathbf{y}_j)$. In this analysis we use patterns defined by \eqref{eq4} and \eqref{eq8}, where $n, m =12$, $\overline{E}_i^*=\overline{E}_i$, and $D_i^*=D_i$.

\begin{figure}
	\centering
	\includegraphics[width=0.7\textwidth]{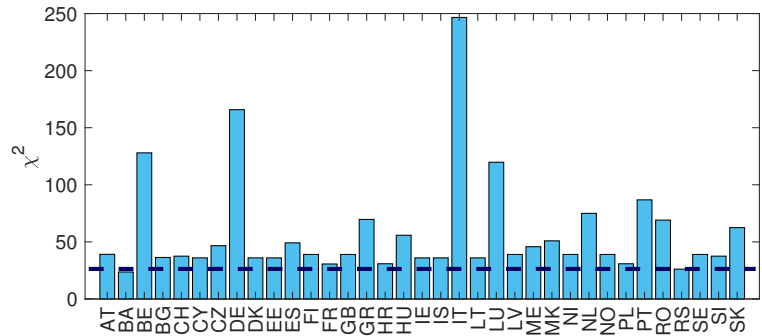}
	\caption{Chi-squared statistics for verification of PSFM assumption.} 
	\label{fig8}
\end{figure}

In the contingency tables both random variables are divided into five categories containing roughly the same number of observations. Fig. \ref{fig8} shows the chi-squared statistics for the analyzed countries. The critical value $\chi^2$ is shown by a dashed line. It is 26.30 for five categories adopted for each random variable and significance level  $\alpha = 0.05$. As you can see from this figure, all $\chi^2$ values are higher or very close to the critical value. This allows us to reject the null hypothesis and justifies the use of PSFMs. The strongest relationships between random variables are observed for Italy, Germany, Belgium and Luxembourg.    

\subsection{Comparative Models}
\label{}

As comparative models we use classical statistical forecasting models such as ARIMA and ETS, as well as neural and neuro-fuzzy models:

\begin{enumerate}
	\item [ARIMA]
	
	-- ARIMA$(p, d, q)(P, D, Q)_{12}$ model implemented in function \texttt{auto.arima} in R environment (package \texttt{forecast}). This function implements automatic ARIMA modeling which combines unit root tests, minimization of the Akaike information criterion (AICc) and maximum likelihood estimation to obtain the optimal ARIMA model \cite{Hyn19}.
	
	\item [ETS] 
	
	-- exponential smoothing state space model \cite{Hyn08} implemented in function \texttt{ets} (R package \texttt{forecast}). This implementation includes many types of ETS models depending on how the seasonal, trend and error components are taken into account. They can be expressed additively or multiplicatively, and the trend can be damped or not. As in the case of \texttt{auto.arima}, \texttt{ets} returns the optimal model estimating the model parameters using AICc \cite{Hyn19}.
	
	\item [MLP] 
	
	-- multilayer perceptron described in \cite{Pel19b}. This model is designed for MTLF. It learns from patterns defined by \eqref{eq4} and \eqref{eq8}. It predicts one component of y-pattern on the basis of x-patterns. For all $m$ components, $m$ MLPs are trained. When all $m$ components of y-pattern are predicted by the set of MLPs, the forecasts of demands are calculated using \eqref{eq9}. The network has one hidden layer with sigmoidal neurons and learns using Levenberg-Marquardt method with Bayesian regularization to prevent overfitting. The MLP hyperparameters which were adjusted are the number of hidden nodes and length of the input patterns $n$. We use Matlab R2018a implementation of MLP (function \texttt{feedforwardnet} from Neural Network Toolbox). 
	
	\item [ANFIS] 
	
	-- adaptive neuro-fuzzy inference system proposed for MTLF in \cite{Pel18}. It works on patterns \eqref{eq4} and \eqref{eq8}. This is an $n$-input, single-output model for prediction of one y-pattern component. So, for all components, $m$ models are built and trained. ANFIS architecture is functionally equivalent to a Sugeno type fuzzy rule base. Initial membership function parameters in the premise parts of rules are determined using fuzzy $c$-means clustering. A hybrid learning method is applied for ANFIS training which uses a combination of the least-squares for consequent parameters and backpropagation gradient descent method for premise parameters. The ANFIS hyperparameters which were selected are the number of rules and the length of the input patterns $n$. The Matlab R2018a implementation of ANFIS was used (function \texttt{anfis} from Fuzzy Logic Toolbox).
	
	\item [LSTM] 
	%https://www.mathworks.com/help/deeplearning/examples/time-series-forecasting-using-deep-learning.html
	-- long short-term memory (LSTM) network, where the responses are the training sequences with values shifted by one time step (a sequence-to-sequence regression LSTM network). For multiple time steps, after one step was predicted the network state was updated. Previous prediction was used as input to the network producing the prediction for the next time step. LSTM was optimized using Adam (adaptive moment estimation) optimizer. The number of hidden nodes was the only hyperparameter to be tuned. Other hyperparameters remain at their default values. The experiments were carried out using Matlab R2018a implementation of LSTM (function \texttt{trainNetwork} from Neural Network Toolbox).

\end{enumerate}

\subsection{Parameter settings and model variants}
\label{} 

Taking into account our earlier experiences with PSFMs reported in \cite{dud15a}, \cite{dud15b}, \cite{Pel18}, \cite{Pel19b}, \cite{Pel19b}, \cite{Pel17} and \cite{Pel17b} we use pattern definitions \eqref{eq4} and \eqref{eq8}. In most cases they provided the best accuracy of the models.      

One of the main hyperparameters for all proposed PSFMs as well as for MLP and ANFIS was the length of the x-patterns. Although the natural choice for this hyperparameter is the seasonal cycle length, i.e. 12, we tested the models in the range for $n$ from 3 to 24, and finally selected the optimal value of $n$ for each model and each time series. 

The $k$-NN model was used in two variants. The first one assigns the same weights for all $k$ neighbors: $w(\mathbf{x}, \mathbf{x}_i)=1/k$. The second one uses weighting function defined by \eqref{eq12} and \eqref{eq13}, where $\rho=1$ and $\gamma=0$ (linear weighting function). This variant is marked as $k$-NNw. The key hyperparameter in these both variants, besides the x-pattern length, is the number of nearest neighbors $k$. It was searched in the range from 1 to 50.      

In FNM we set $\alpha=2$ and control the membership function \eqref{eq14} width with $\sigma$. This hyperparameter was calculated from: 

\begin{equation}\label{eq20}
\sigma = a \cdot d_{med}
\end{equation}
where $d_{med}$ was the median of pairwise distances between x-patterns in the training set. It was searched for $a = 0.02, 0.04, …, 1$. 

Determining $\sigma$ on the basis of $d_{med}$ calibrates this parameter to data.    

The bandwidth parameters in N-WE were searched around the starting values determined using the Scott rule \cite{Scott} proposed for the normal product density estimators:

\begin{equation}\label{eq21}
h^S_t=s_t N^{-\frac{1}{n+4}},\quad t=1,2,...,n
\end{equation}
where $s_t$ is the standard deviation of the $t$-th component of $\mathbf{x}$ estimated from the training sample.

The searched bandwidths are generated according to:
\begin{equation}\label{eq22}
h_t=b \cdot h^S_t,\quad t=1,2,...,n
\end{equation}
where $b = 0.15, 0.2, …, 2$.

Note that the multidimensional optimization problem of searching bandwidths $h_1, h_2, ..., h_n$ was replaced by a simple one-dimensional problem of searching $b$.  

For GRNN we assume in this study the same bandwidths for all neurons. The bandwidth $\sigma$ was searched according to \eqref{eq20}. For MLP the number of hidden neurons was selected from the range from 1 to 10. The number of rules in the ANFIS model was selected from the range from 2 to 13. The number of hidden nodes in the LSTM model was selected from the set $\{1, 2, ..., 10, 15, ...50, 60, ..., 200\}$. For ARIMA and ETS models we use default parameter settings implemented in functions \texttt{auto.arima} and \texttt{ets}, respectively.     

The optimal values of the hyperparameters for each model and for each of 35 time series were selected in the grid search procedure using cross-validation.  

Taking into account the y-pattern encoding variants described in Section 2, three variants of each PSFM as well as MLP and ANFIS models are considered:
\begin{enumerate}
	\item The basic variant, where the coding variables for y-patterns are the mean and dispersion of sequence $X_i$, i.e. $\overline{E}_i^*=\overline{E}_i$, $D_i^*=D_i$. This variant enables us to calculate the forecast of the monthly loads from \eqref{eq9} without additional forecasting the coding variables. 
	\item The variant, were coding variables are the mean and dispersion of sequence $Y_i$. They are both forecasted independently using ARIMA model on the basis of their historical values. The symbols of models in this variant are extended by "+ARIMA", e.g. "$k$-NN+ARIMA", "ANFIS+ARIMA".
	\item As in variant 2, the coding variables are the mean and dispersion of sequence $Y_i$. But in this case they are forecasted using ETS. The symbols of models in this variant are extended by "+ETS", e.g. "$k$-NN+ETS", "ANFIS+ETS".
\end{enumerate} 

\subsection{Results and analysis}
\label{} 

The PSFMs are deterministic models which return the same results for the same data. NN-based models, i.e. MLP, ANFIS and LSTM, due to the stochastic nature of the learning processes return different results for the same data. In this study these models were trained 100 times and the final errors were calculated as the averages over 100 independent trials.

\begin{figure}
	\centering
	\includegraphics[width=1\textwidth]{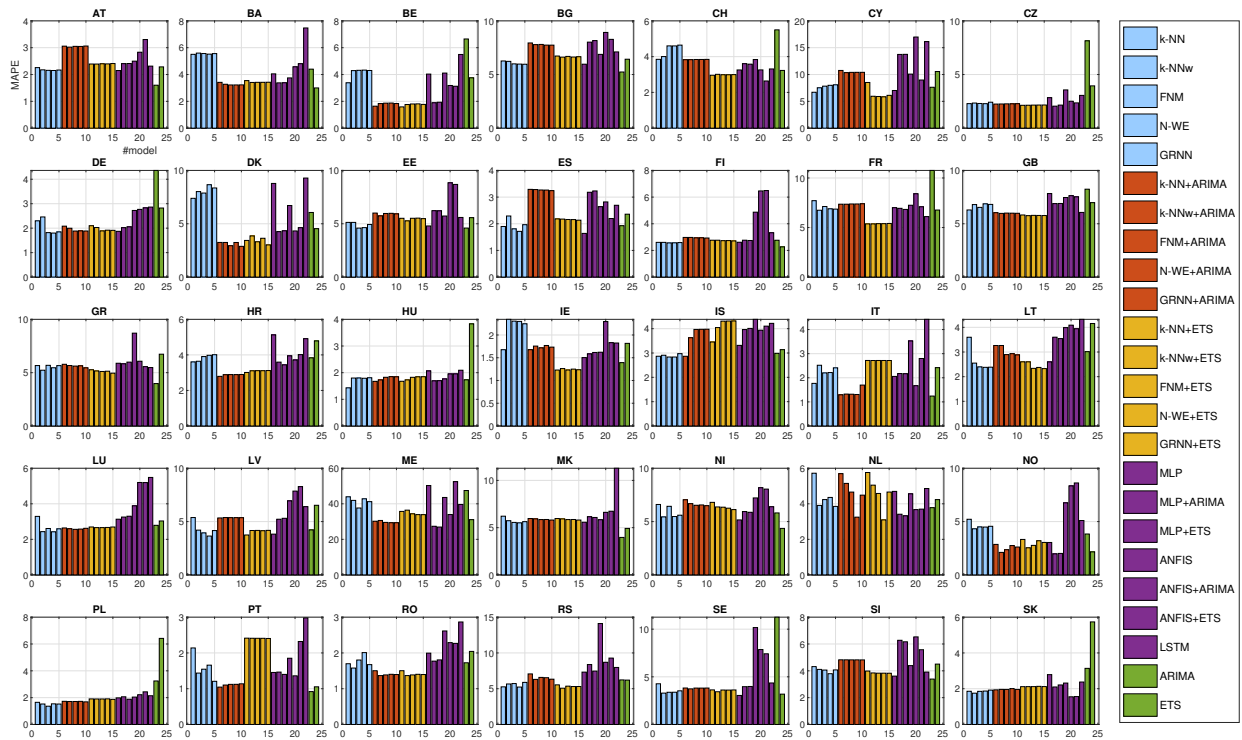}
	\caption{MAPE for each country.} 
	\label{fig9}
\end{figure}

\begin{figure}
	\centering
	\includegraphics[width=0.49\textwidth]{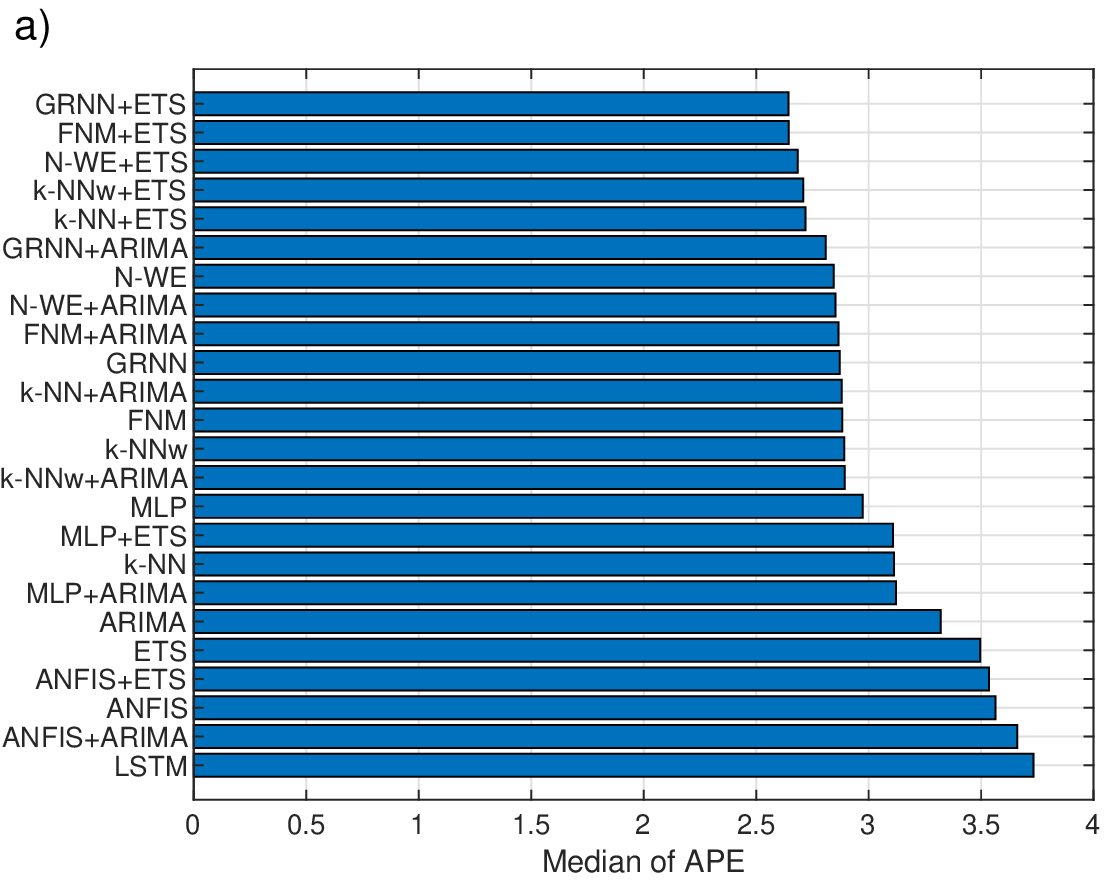}
	\includegraphics[width=0.49\textwidth]{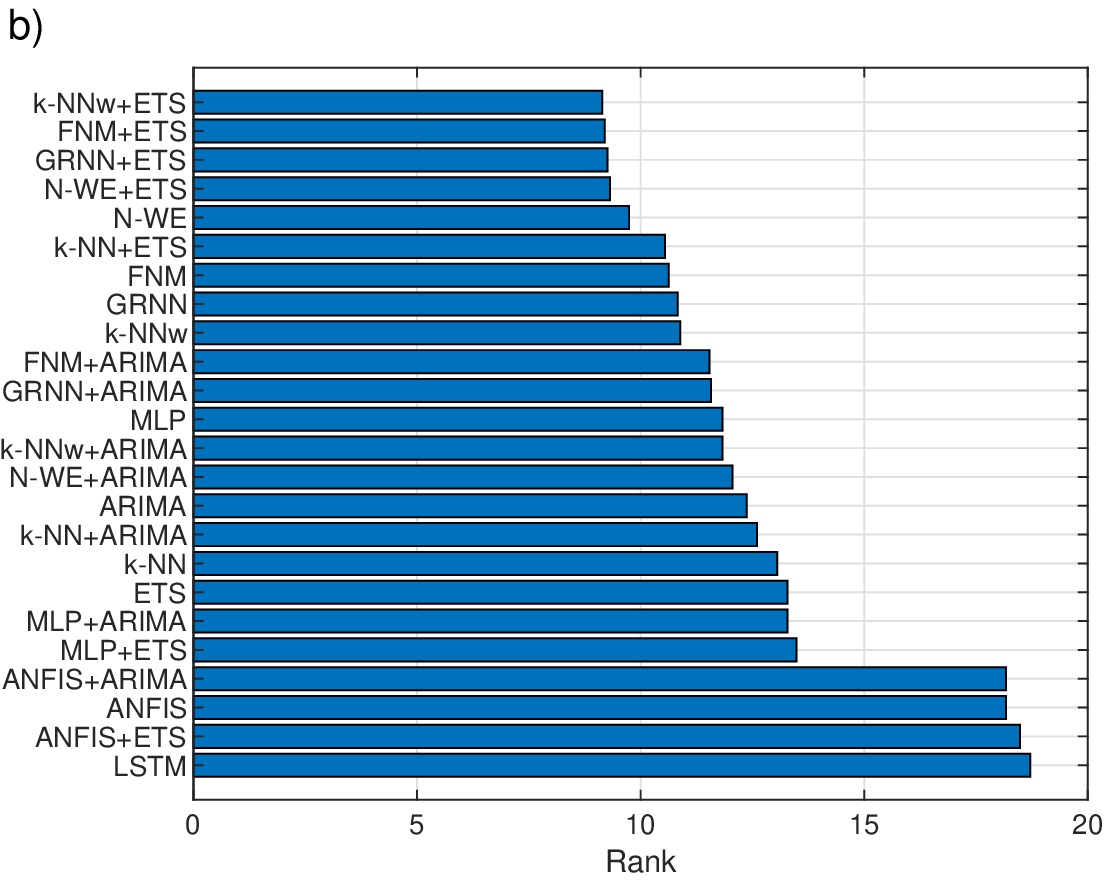}
	\caption{Rankings of the models: (a) ranking based on the median of APE, (b) ranking based on the average model ranks in the individual rankings for each country.} 
	\label{fig10}
\end{figure}

Fig. \ref{fig9} shows the forecasting errors on the test sets (mean absolute percentage error, MAPE) for each country. The rankings of the models are shown in Fig. \ref{fig10}. The ranking shown on the left is based on the median of APE and the ranking shown on the right is based on the average ranks of the models in the rankings for individual countries. Table 1 summarizes the accuracy of the models showing median of APE, MAPE, interquartile ranges of APE averaged over all countries and root mean square error (RMSE) also averaged over all countries.

\begin{table}[]
	\caption{Results comparison among proposed and comparative models.}\label{tab1}
	\setlength{\tabcolsep}{10pt}
	\centering
	\begin{tabular}{lrrrr}
		\hline
		Model       & \multicolumn{1}{c}{Median \textit{APE}} & \multicolumn{1}{c}{\textit{MAPE}} & \multicolumn{1}{c}{\textit{IQR}} & \multicolumn{1}{c}{\textit{RMSE}} \\ \hline
		k-NN        & 3.11                                    & 5.19                              & 4.13                             & 385.68                            \\
		k-NNw       & 2.89                                    & 4.99                              & 4.06                             & 368.79                            \\
		FNM         & 2.88                                    & 4.88                              & 4.43                             & 354.33                            \\
		N-WE        & 2.84                                    & 5.00                              & 4.14                             & 352.01                            \\
		GRNN        & 2.87                                    & 5.01                              & 4.30                             & 350.61                            \\
		k-NN+ARIMA  & 2.88                                    & 4.71                              & 4.21                             & 352.42                            \\
		k-NNw+ARIMA & 2.89                                    & 4.65                              & 4.02                             & 346.58                            \\
		FNM+ARIMA   & 2.87                                    & 4.61                              & 3.83                             & 341.41                            \\
		N-WE+ARIMA  & 2.85                                    & 4.59                              & 3.74                             & 340.26                            \\
		GRNN+ARIMA  & 2.81                                    & 4.60                              & 3.77                             & 345.46                            \\
		k-NN+ETS    & 2.72                                    & 4.58                              & 3.55                             & 333.27                            \\
		k-NNw+ETS   & 2.71                                    & 4.47                              & 3.43                             & 327.94                            \\
		FNM+ETS     & 2.64                                    & 4.40                              & 3.34                             & 321.98                            \\
		N-WE+ETS    & 2.68                                    & 4.37                              & 3.20                             & 320.51                            \\
		GRNN+ETS    & 2.64                                    & 4.38                              & 3.35                             & 324.91                            \\
		MLP         & 2.97                                    & 5.27                              & 3.89                             & 378.81                            \\
		MLP+ARIMA   & 3.12                                    & 4.83                              & 4.16                             & 362.03                            \\
		MLP+ETS     & 3.11                                    & 4.80                              & 4.12                             & 358.07                            \\
		ANFIS       & 3.56                                    & 6.18                              & 4.91                             & 488.75                            \\
		ANFIS+ARIMA & 3.66                                    & 6.05                              & 5.40                             & 473.80                            \\
		ANFIS+ETS   & 3.54                                    & 6.32                              & 4.57                             & 464.29                            \\
		LSTM        & 3.73                                    & 6.11                              & 4.46                             & 431.83                            \\
		ARIMA       & 3.32                                    & 5.65                              & 5.27                             & 463.07                            \\
		ETS         & 3.50                                    & 5.05                              & 4.17                             & 374.52                            \\ \hline
	\end{tabular}
\end{table}

As can be seen from Fig. \ref{fig10} and Table 1, the most accurate models are PSFMs with ETS for forecasting the coding variables. There is no significant difference in errors between them. PSFMs in the basic variants and in +ARIMA variants were ranked lower in both rankings then PSFMs+ETS. The largest errors among PSFMs were observed for the simple $k$-NN model with equal weights. The comparative models were ranked as the least accurate. Among them the best one turned out to be MLP and the worse ones were ANFIS-based models and LSTM  ($MAPE > 6\%$).  

Fig. \ref{fig9} allow us to compare the variants of PSFMs in more detail. The basic variants use the coding variables determined from history, i.e. the means and dispersions of sequence $X_i$. In such a case, the stability of relationships between means and dispersions, respectively, of sequences $X_i$ and $Y_i$ is very important. When the trend is falling the y-patterns are encoded with the mean value of the previous sequence $X_i$ which is higher then the mean of $Y_i$. For a forecasted y-pattern the same higher value of the mean coding variable is expected. But when the time series instead of keeping falling starts to rise, the relationship between means of sequence $X_i$ and $Y_i$ observed in the history is no longer valid. This results in a wrong forecast which continues a falling trend. The similar problem arises when the trend is rising and it starts to fall in a final part. This problem is observed for many countries, e.g. BA, BE, IT, DK, IE and others (see the increased errors for PSFMs in the basic variants for these countries in Fig. \ref{fig9}). To prevent this situation the coding variables can be forecasted for the sequence $Y_i$. We use ARIMA and ETS for this. In this case, the final accuracy is dependent on the accuracies of the three models: PSFM which predicts y-pattern, ARIMA or ETS which predict the mean demand of sequence $Y_i$, and ARIMA or ETS which predict dispersion of sequence $Y_i$. When the coding variables are predicted with low accuracy, the final error can be higher than in the case of the basic PSFMs (see graphs for AT, BG and ES in Fig. \ref{fig9}). Note the enormous errors for ME, $MAPE > 25\%$. They are caused by a very irregular character of the time series for ME and abnormal value of demand for March and April 2013, which is about twice the value typical for these months. No model has managed this time series sufficiently.             

\begin{figure}
	\centering
	\includegraphics[width=0.435\textwidth]{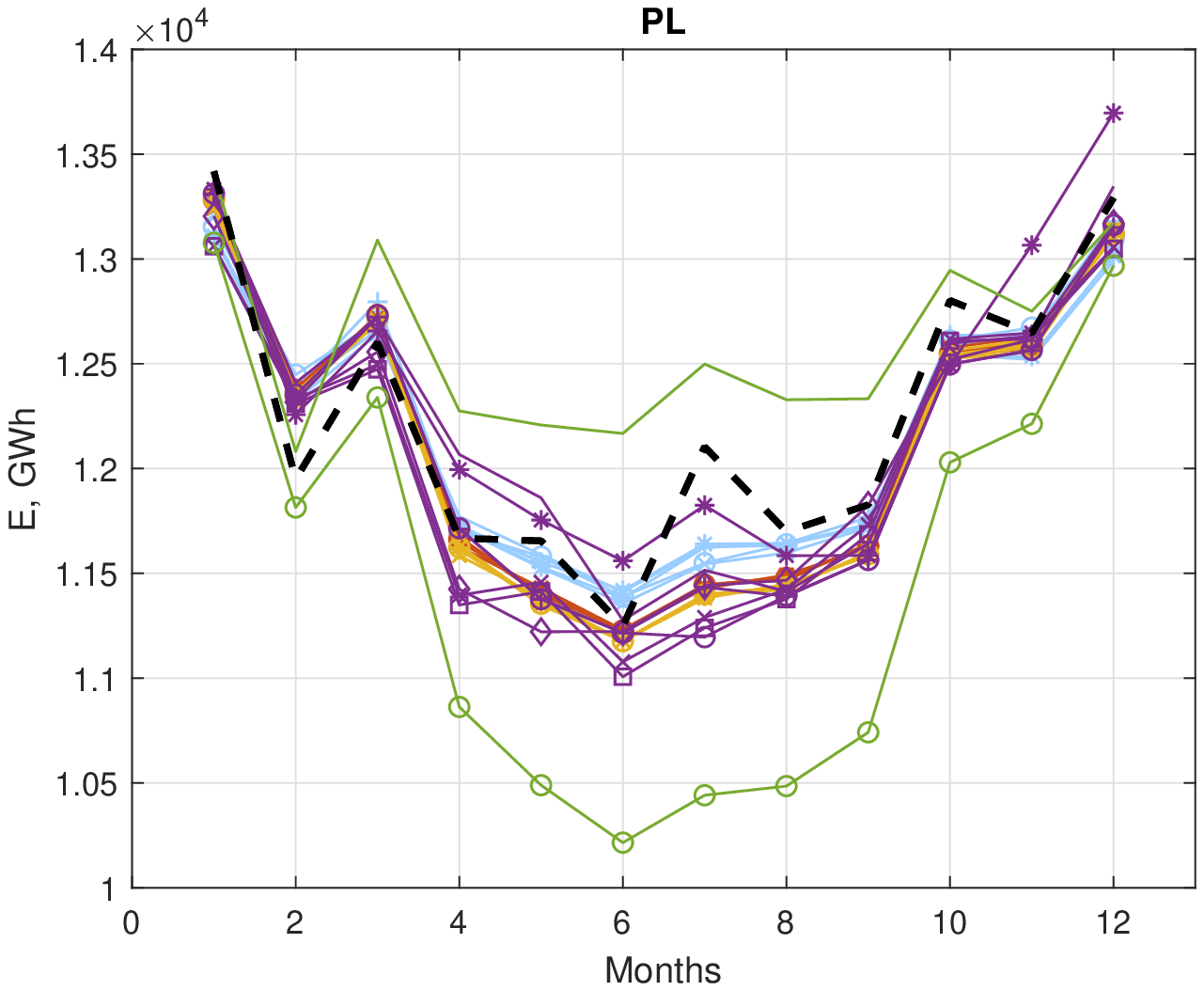}
	\includegraphics[width=0.435\textwidth]{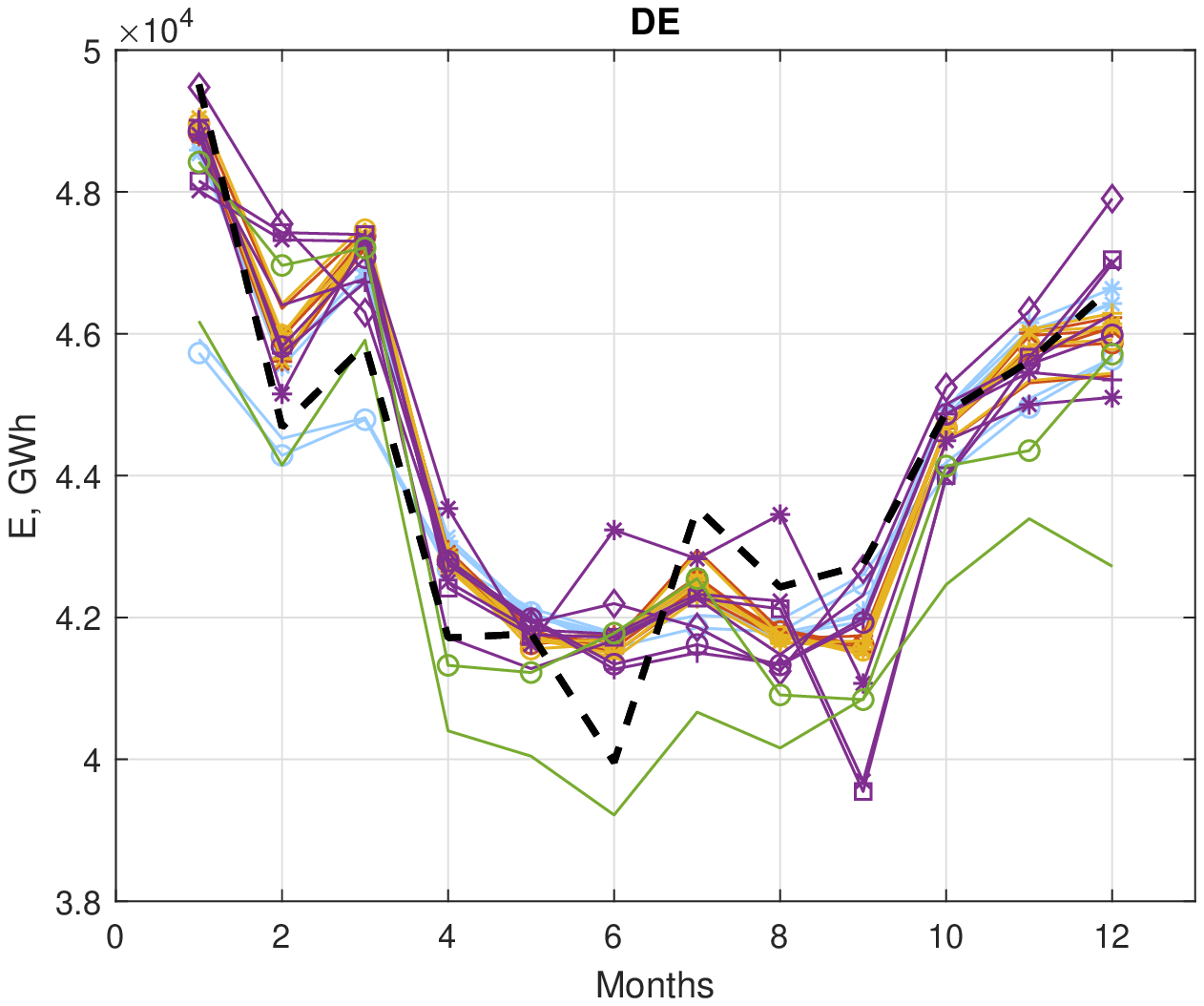}
	\includegraphics[width=0.11\textwidth]{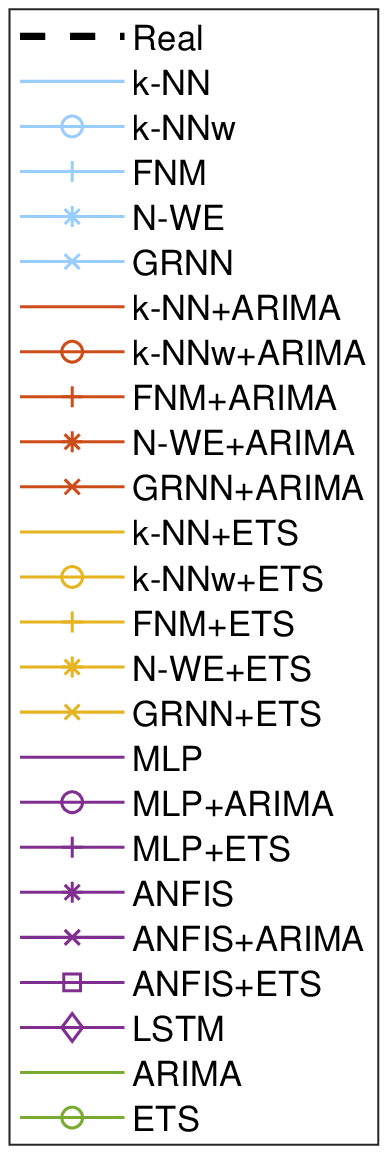}
	\includegraphics[width=0.435\textwidth]{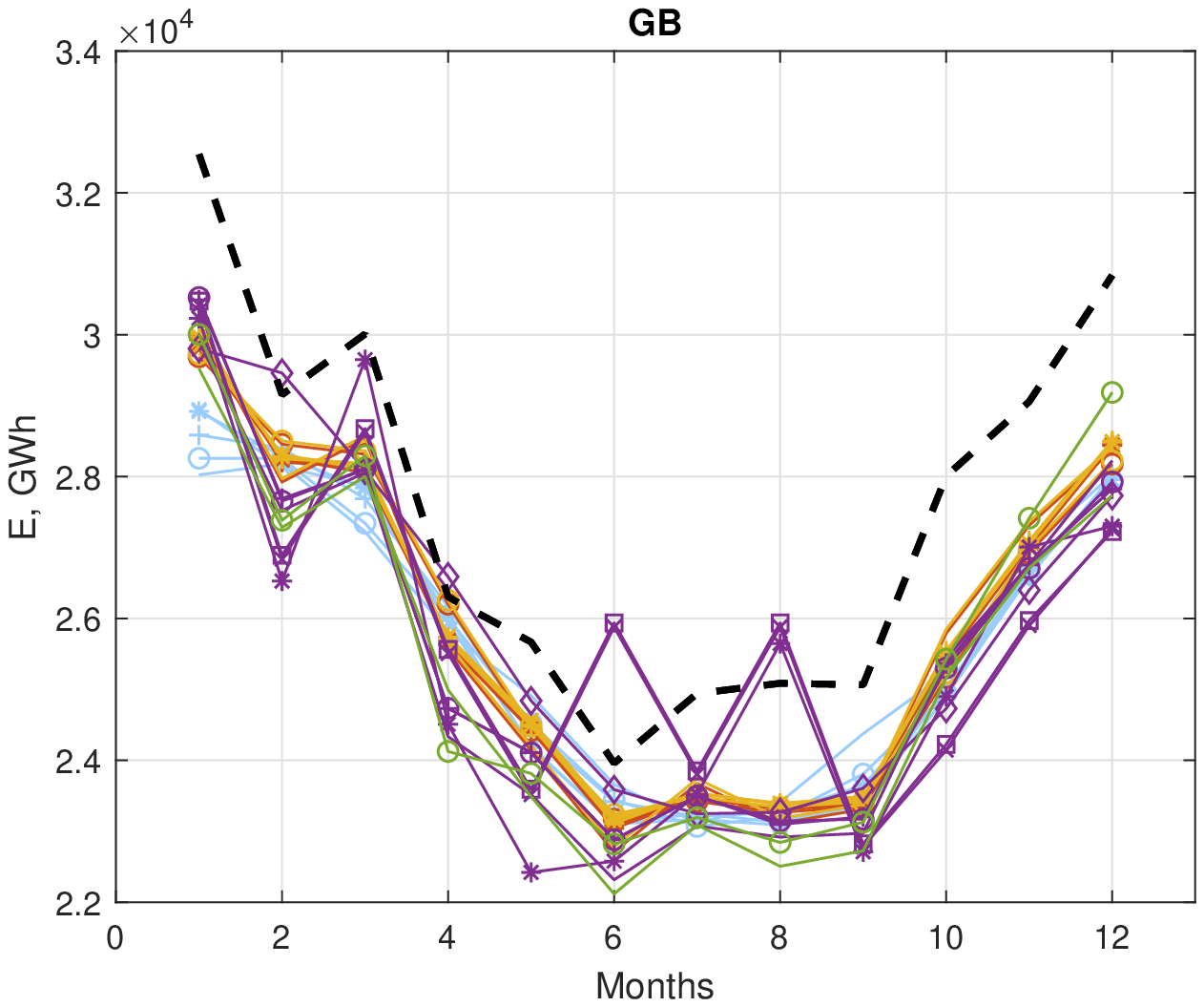}
	\includegraphics[width=0.435\textwidth]{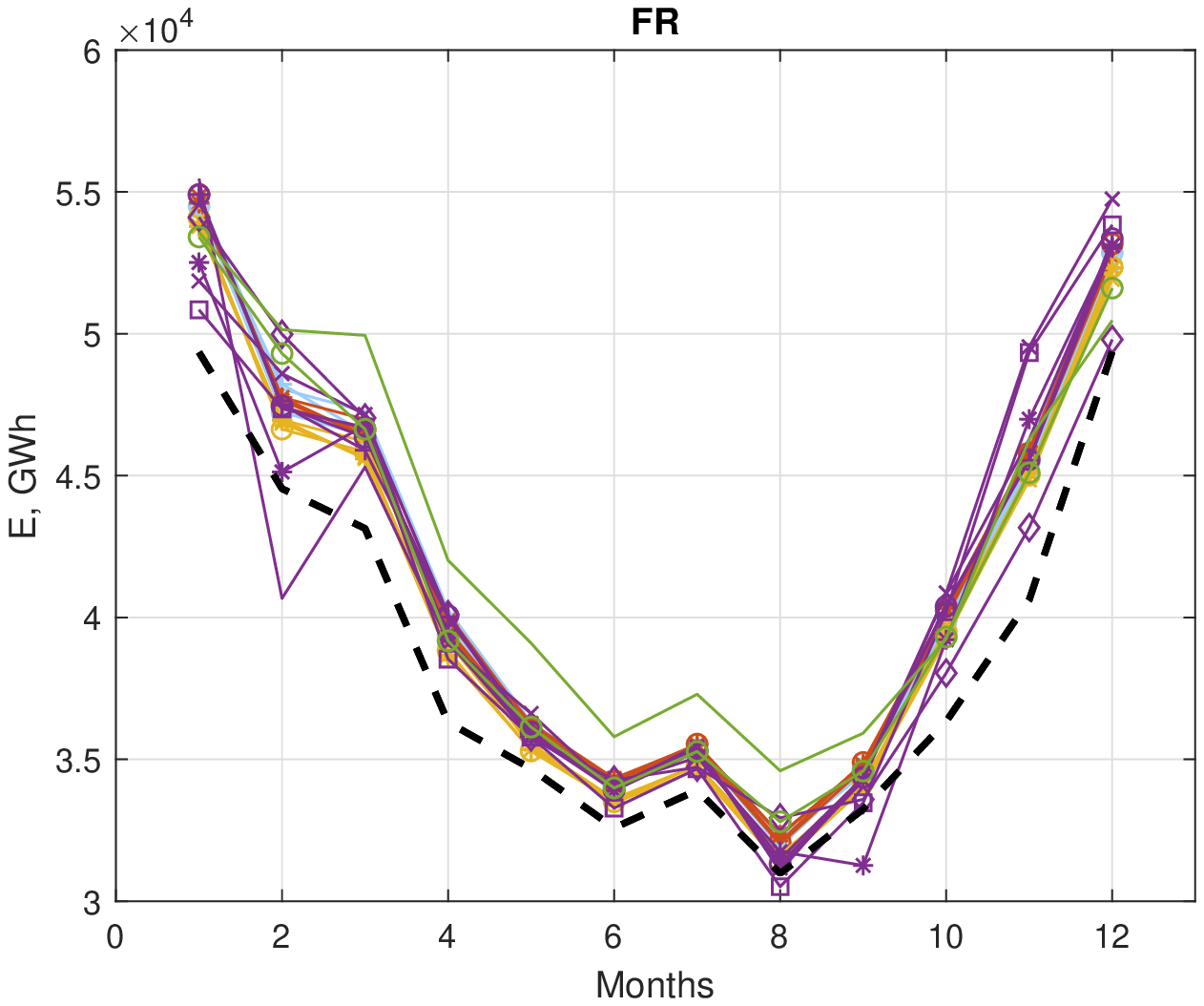}
	\includegraphics[width=0.11\textwidth]{Prog_pusty.eps}
	\caption{Real and forecasted electricity demand for PL, DE, GB and FR.} 
	\label{fig11}
\end{figure}

In Fig. \ref{fig11} examples of forecasts generated by the models for four countries are depicted. For PL the PSFMs produce the most accurate forecasts. Among PSFMs the basic variant is slightly better than +ARIMA and +EST variants. Note that for PL the classical models, ARIMA nad ETS, give outlier forecasts. ARIMA produces overestimated forecasts while ETS produces underestimated forecasts. ARIMA also gives most inaccurate forecasts for DE and FR. For GB the forecasts generated by all models are underestimated. It results from that demand in 2014 went up unexpectedly despite the downward trend observed from 2010 to 2013. The opposite situation for FR caused a slight overestimation of forecasts. Note that for MLP and ANFIS the jumps in the forecasted curve are observed. They results from that these models predict only one component of the y-pattern, and to forecast all $m$ components we use $m$ independent models. So, the relationships between components are ignored. In the case of PSFMs which generate multi-output response, these relationships are kept because the forecasted annual cycle is formed by weighted averaging the shapes of historical annual cycles.

\section{Conclusion}

In this work we explored pattern similarity-based models for mid-term load forecasting. The key component of these models is the time series representation by patterns of the time series sequences. We defined input and output patterns which unify input and output data. Patterns carry information about the shapes of the annual cycles filtering out the trend and normalizing data. Pattern representation of the time series simplifies the relationships between data. As a result the forecasting model can be simpler and faster in learning.    

The PSFMs belong to the class of lazy learning regression models where the regression function is built by aggregation of the output patterns with weights dependent on the similarity between input patterns. The big advantage of PSFMs are the small number of parameters to adjust. One of them is the input pattern length and another one is the bandwidth parameter deciding about the weighting function shape. The simplest PSFM is the $k$-NN model with equal weights for all $k$ nearest neighbors of the query pattern. More sophisticated models using weighting functions, such as $k$-NNw, FNM, N-WE and GRNN, produce more accurate forecasts. In the variants used in this work, they all have only one hyperparameter controlling the bandwidth of the weighting function and hence, the bias-variance tradeoff of the model. Note that small number of parameters increases the generalization ability and facilitates the model optimization. Another advantage o PSFMs is no need to re-train the model when new data arrives. New data can be immediately added to the training set and are available for the model to produce the forecast. Note that the PSFM generate a vector as an output. Its size, as well as the size of the input pattern, does not affect the number of parameters and training process complexity such as in the case of MLP and ANFIS. The advantages of PSFMs include also robustness to incomplete input information. The models can work using x-patterns with lacking components.  

The simulation study has shown high accuracy of PSFMs when comparing to the classical models such as ARIMA and ETS as well as MLP, ANFIS and LSTM models. The best performance was achieved by the combined models, PSFMs with ETS for forecasting the coding variables. But also basic PSFMs outperform the comparative models generating more accurate forecasts and being much more simpler and easier to optimize.  

The directions of further research on PSFMs are: introduction additional input variables to the models, ensembles of the models, introduction confidence degrees of the training data to the models, and probabilistic forecasting.

%\bibliographystyle{unsrt}
%\bibliography{references}  %%% Remove comment to use the external .bib file (using bibtex).
%%% and comment out the ``thebibliography'' section.

%%% Comment out this section when you \bibliography{references} is enabled.

\begin{thebibliography}{}
	
	\bibitem{Ahm19}
	T. Ahmad, H. Chen, Potential of three variant machine-learning models for forecasting district level medium-term and long-term energy demand in smart grid environment, Energy 160 (2018) 1008-1020.
	
	\bibitem{AlH05}
	H.M. Al-Hamadi, S.A. Soliman, Long-term/mid-term electric load forecasting based on short-term correlation and annual growth, Electric Power Syst. Res. 74 (2005) 353–361.
	
	\bibitem{Apa12}
	F. Apadula, A. Bassini, A. Elli, S. Scapin, Relationships between meteorological variables and monthly electricity demand. Appl Energy 98 (2012) 346–356.% doi: 10.1016/j.apenergy.2012.03.053
	
	\bibitem{Atk97}
	C.G. Atkenson, A.W. Moor, S. Schaal, Locally weighted learning. Artificial Intelligence Review 11 (1997) 75–113.
	
	\bibitem{Bar01}
	E.H. Barakat, Modeling of nonstationary time-series data. Part II. Dynamic periodic trends. Electr Power Energy Systems 23 (2001) 63–68.
	
	\bibitem{Bed18}
	J. Bedi, D. Toshniwal, Empirical mode decomposition based deep learning for electricity demand forecasting,  IEEE Access 6 (2018) 49144-49156.
	
	\bibitem{Ben06}
	D. Benaouda, F. Murtagh, J.L. Starck, O. Renaud, Wavelet-based nonlinear multiscale decomposition model for electricity load forecasting. Neurocomputing 70(1–3) (2006) 139–154.% doi:10.1016/j.neucom.2006.04.005
	
	\bibitem{Bun09}
	P. Bunnoon, K. Chalermyanont, C. Limsakul, Mid term load forecasting of the country using statistical methodology: Case study in Thailand, International Conference on Signal Processing Systems (2009) 924-928.
	
	\bibitem{Pei11}
	Chang Pei-Chann, Fan Chin-Yuan, Lin Jyun-Jie, Monthly electricity demand forecasting based on a weighted evolving fuzzy neural network approach,  Electrical Power and Energy Systems 33 (2011), 17–27.
	
	\bibitem{chen09}
	Y. Chen, E.K. Garcia, M. R. Gupta, A. Rahimi, L. Cazzanti, Similarity-based classification: Concepts and algorithms, Journal of Machine Learning Research 10 (2009) 747-776.
	
	\bibitem{Chen17}
	J.F. Chen, S.K. Lo, Q.H. Do, Forecasting monthly electricity demands: An application of neural networks trained by heuristic algorithms, Information 8(1) (2017) 31.
	
	\bibitem{Oli18}
	E.M de Oliveira, F.L.C. Oliveira, Forecasting mid-long term electric energy consumption through bagging ARIMA and exponential smoothing methods, Energy 144 (2018) 776-788
	
	\bibitem{Dog16}
	E. Dogan, Are shocks to electricity consumption transitory or permanent? Sub-national evidence from Turkey, Utilities Policy 41 (2016) 77–84. %doi: 10.1016/j.jup.2016.06.007
	
	\bibitem{Dov99}
	E. Doveh, P. Feigin., L. Hyams, Experience with FNN models for medium term power demand predictions, IEEE Trans. Power Syst. 14 (2) (1999) 538-546.
	
	\bibitem{duch}
	W. Duch, Similarity-based methods: A general framework for classification, approximation and association, Control and Cybernetics 29 (4) (2000) 937-968.
	
	\bibitem{dud15b}
	G. Dudek, Pattern similarity-based methods for short-term load forecasting – Part 2: Models, Applied Soft Computing 36 (2015) 422-441.
	
	\bibitem{dud15a}
	G. Dudek, Pattern similarity-based methods for short-term load forecasting – Part 1: Principles, Applied Soft Computing 37 (2015) 277-287.
	
	\bibitem{Pel17b}
	G. Dudek, P. Pełka, Medium-term electric energy demand forecasting using Nadaraya-Watson estimator. Proc. Conf. on Electric Power Engineering EPE'17 (2017) 1-6.
	
	\bibitem{Elk98}
	M.M. Elkateb, K. Solaiman, Y. Al-Turki, A comparative study of medium-weather-dependent load forecasting using enhanced artificial/fuzzy neural network and statistical techniques, Neurocomputing 23 (1998) 3–13.
	
	\bibitem{Gav01}
	M. Gavrilas, I. Ciutea, C. Tanasa, Medium-term load forecasting with artificial neural network models, IEEE Conf. Elec. Dist. Pub. 6 (2001).
	
	\bibitem{Ghi06}
	M. Ghiassi, D.K. Zimbra, H. Saidane, Medium term system load forecasting with a dynamic artificial neural network model, Electric Power Systems Research 76 (2006) 302–316
	
	\bibitem{Gon06}
	E. González-Romera, M.A. Jaramillo-Morán, D. Carmona-Fernández, Monthly electric energy demand forecasting based on trend extraction. IEEE Trans Power System 21(4) (2006) 1935–1946.
	
	\bibitem{Gon08}
	E. González-Romera, M.A. Jaramillo-Morán, D. Carmona-Fernández, Monthly electric energy demand forecasting with neural networks and Fourier series, Energy Conversion and Management 49 (2008) 3135–3142.
	
	\bibitem{Har04}
	W.K. Härdle, M. Müller, S. Sperlich, A. Werwatz, Nonparametric and Semiparametric Models. Springer, 2004.
	
	\bibitem{Hyn19}
	R.J. Hyndman, G. Athanasopoulos, Forecasting: principles and practice, 2nd edition, OTexts: Melbourne, Australia. OTexts.com/fpp2 (2018) Accessed on 4 October 2019.
	
	\bibitem{Hyn08}
	R.J. Hyndman, A.B. Koehler, J.K. Ord, R.D. Snyder, Forecasting with exponential smoothing: The state space approach, Springer, 2008.
	
	\bibitem{Kan02}
	M.S. Kandil, S.M. El-Debeiky, N.E. Hasanien,Long-term load forecasting for fast developing utility using a knowledge-based expert system, IEEE Trans. Power Syst. 17(2) (2002) 491–496.
	
	\bibitem{Moh18}
	N.A. Mohammed, Modelling of unsuppressed electrical demand forecasting in Iraq for long term, Energy 162 (2018) 354-363.
	
	\bibitem{Pel19}
	P. Pełka, G. Dudek, Medium-term electric energy demand forecasting using generalized regression neural network. Proc. Conf.  Information Systems Architecture and Technology ISAT 2018, AISC 853, Springer, Cham (2018) 218-227. 
	
	\bibitem{Pel18}
	P. Pełka, G. Dudek, Neuro-fuzzy system for medium-term electric energy demand forecasting, Proc. Conf.  Information Systems Architecture and Technology ISAT 2017, AISC 655, Springer, Cham (2018) 38-47.
	
	\bibitem{Pel19b}
	P. Pełka, G. Dudek, Pattern-based forecasting monthly electricity demand using multilayer perceptron, Proc. Conf. Artificial Intelligence and Soft Computing ICAISC 2019, LNAI 11508, Springer, Cham (2019) 663-672. 
	
	\bibitem{Pel17}
	P. Pełka, G. Dudek, Prediction of monthly electric energy consumption using pattern-based fuzzy nearest neighbour regression. Proc. Conf. Computational Methods in Engineering Science CMES'17, ITM Web Conf.15 (2017) 1-5.
	
	\bibitem{Qiu17}
	X. Qiu, Y. Ren, P.N. Suganthan, G.A.J. Amaratunga, Empirical mode decomposition based ensemble deep learning for load demand time series forecasting. Appl Soft Comp 54 (2017) 246–255. %doi:10.1016/j.asoc.2017.01.015
	
	\bibitem{Scott}
	D.W. Scott, Multivariate density estimation: Theory, practice, and visualization. Wiley, 1992.
	
	\bibitem{Spe91}
	D.F. Specht, A general regression neural network. IEEE Trans Neural Networks 2(6) (1991) 568–576.
	
	\bibitem{Sug11}
	L. Suganthi, A.A. Samuel, Energy models for demand forecasting — A review. Renew Sust Energy Rev 16(2) (2012) 1223–1240. %doi: 10.1016/j.rser.2011.08.014
	
	\bibitem{The11}
	M. Theodosiou, Forecasting monthly and quarterly time series using STL decomposition. Int J Forecast 27(4) (2011) 1178–1195. %doi: 10.1016/j.ijforecast.2010.11.002
	
	\bibitem{Zhao12}
	W. Zhao, F. Wang, D. Niu, The application of support vector machine in load forecasting, Journal of Computers 7(7) (2012) 1615-1622.

\end{thebibliography}

\end{document}